\pdfoutput=1

\documentclass[11pt]{article}
\usepackage{algorithm}
\usepackage{amssymb}
\usepackage{algpseudocode}
\usepackage{subfigure} 
\usepackage{multirow}
\usepackage{booktabs}
\usepackage{graphicx}
\usepackage{geometry}
\usepackage{array}
\usepackage{amsmath}
\usepackage[table]{xcolor}
\usepackage{bm}
\usepackage[preprint]{acl}

\usepackage{times}
\usepackage{latexsym}

\usepackage[T1]{fontenc}

\usepackage[utf8]{inputenc}

\usepackage{microtype}
\usepackage{booktabs}
\usepackage{tabularx}
\usepackage{inconsolata}

\usepackage{graphicx}

\title{Training LLMs to be Better Text Embedders through \\ Bidirectional Reconstruction}

\author{
  \textbf{Chang Su\textsuperscript{1}}, 
  \textbf{Dengliang Shi\textsuperscript{2}}, 
  \textbf{Siyuan Huang\textsuperscript{1}}, 
  \textbf{Jintao Du\textsuperscript{2}}, 
  \textbf{Changhua Meng\textsuperscript{2}}, \\
  \textbf{Yu Cheng\textsuperscript{2}}, 
  \textbf{Weiqiang Wang\textsuperscript{2}}, 
  \textbf{Zhouhan Lin\textsuperscript{3,\thanks{Corresponding Author.}}}
\\
  \textsuperscript{1}LUMIA Lab, Shanghai Jiao Tong University \\
  \textsuperscript{2}Tiansuan Lab, Ant Group Co., Ltd. \\
  \textsuperscript{3}LUMIA Lab, School of Artificial Intelligence, Shanghai Jiao Tong University
\\
  \texttt{suchang0912@sjtu.edu.cn, siyuan\_huang\_sjtu@outlook.com, lin.zhouhan@gmail.com}\\
  \texttt{\{dengliang.sdl,lingke.djt,changhua.mch,cy122623,weiqiang.wwq\}@antgroup.com}
}

\begin{document}
\maketitle
\begin{abstract}
Large language models (LLMs) have increasingly been explored as powerful text embedders. Existing LLM-based text embedding approaches often leverage the embedding of the final token, typically a reserved special token such as \texttt{[EOS]}. However, these tokens have not been intentionally trained to capture the semantics of the whole context, limiting their capacity as text embeddings, especially for retrieval and re-ranking tasks. 
We propose to add a new training stage before contrastive learning to enrich the semantics of the final token embedding. This stage employs bidirectional generative reconstruction tasks, namely EBQ2D (Embedding-Based Query-to-Document) and EBD2Q (Embedding-Based Document-to-Query), which interleave to anchor the \texttt{[EOS]} embedding and reconstruct either side of Query-Document pairs. 
Experimental results demonstrate that our additional training stage significantly improves LLM performance on the Massive Text Embedding Benchmark (MTEB), achieving new state-of-the-art results across different LLM base models and scales.\footnote{Our code is available at \url{https://github.com/LUMIA-Group/Anchor-Embedding}.}  
\end{abstract}

\section{Introduction}

\renewcommand{\dblfloatpagefraction}{.9}
\begin{figure*}[htbp]
\centering
\includegraphics[width=\textwidth]{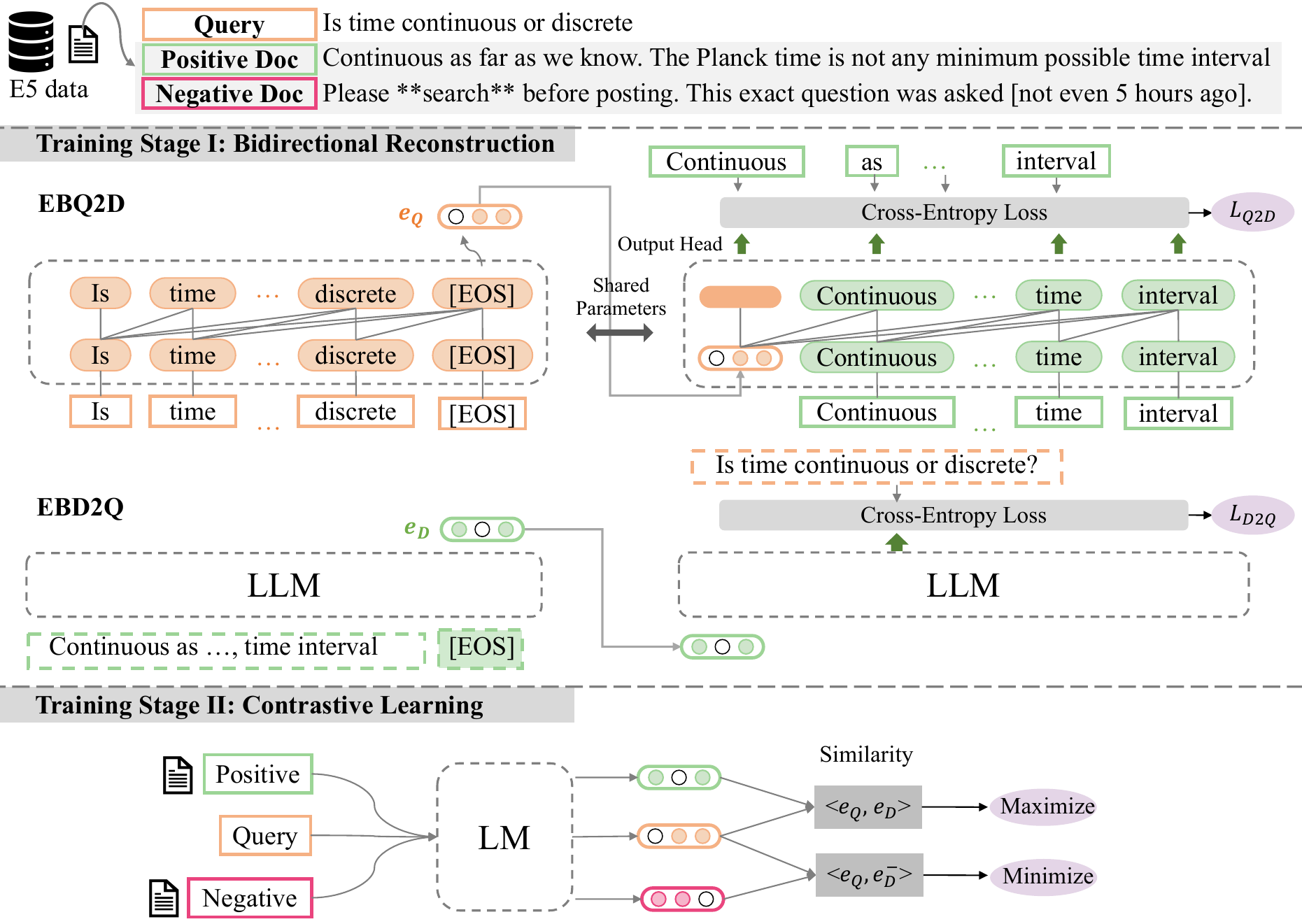}
\caption{The pipeline of our approach. The model is first trained using two bidirectional reconstruction tasks, followed by contrastive learning. The public E5 data serves as the training corpus for both stages.}

\label{fig:pipeline}
\end{figure*}

Text embeddings serve as the foundation for many natural language processing (NLP) tasks by capturing the semantic meaning of text in vector representations \citep{muennighoff2023mteb,lewis2020retrieval}. For example, in text retrieval, both queries and documents are encoded into a shared latent space, where their relevance is measured by embedding similarity, which in turn places strong demands on embedding quality \citep{karpukhin2020dense}.

Early studies leveraged pre-trained language models with bidirectional attention, such as BERT \citep{devlin2019bert} and T5 \citep{raffel2020exploring}, to generate high-quality text embeddings. These approaches typically relied on complex multi-stage training and large-scale annotated pairs \citep{wang2022text,xiao2024c}.

More recently, the impressive semantic understanding capability of large language models (LLMs) has attracted growing interest in their use for embedding tasks. Some approaches transform LLMs into text encoders by enabling bidirectional attention \citep{behnamghader2024llm2vec,muennighoff2024generative}, but such architectural modifications compromise the unification between generation and embedding. Alternatively, other methods retain the auto-regressive nature and causal attention, deriving embeddings from the final token (e.g., \texttt{<\textbackslash s>} or \texttt{[EOS]}) to capture global context, a practice  that has been widely adopted \citep{li2025making, springer2025repetition}. However, during general pre-training, these tokens serve merely as sequence delimiters, and the model does not learn to encode contextual semantics into their representations or to establish meaningful alignment between relevant texts with them. This greatly limits the potential of LLMs in embedding tasks.



Motivated by this, we propose a new training stage before contrastive learning, establishing a two-stage training framework as illustrated in Figure~\ref{fig:pipeline}. For the new stage, we introduce two bidirectional reconstruction tasks, \textbf{EBQ2D} (Embedding-Based Query-to-Document) and \textbf{EBD2Q} (Embedding-Based Document-to-Query), which treat the \texttt{[EOS]} embedding as an anchor to: 1) aggregate the semantic information of either the query or the relevant document, and 2) serve as the reference for generating its counterpart. Specifically, in the EBQ2D task, the output embedding of the \texttt{[EOS]} token in the query is used to prompt the model to generate the relevant document. This encourages the model to embed the semantics of the query and, more importantly, the implied document-level content within the embedding. Symmetrically, the EBD2Q task uses the \texttt{[EOS]} embedding of the document to guide the generation of the corresponding query, training the model to reason backward from content to intent. This training stage enhances the model's ability to capture implicit semantic relationships between queries and relevant documents via the output embedding. Following the bidirectional reconstruction, the second stage fine-tunes the model with contrastive learning to further improve the quality of the generated representations.

We apply our framework to several decoder-only LLMs, including LLaMA-3.1, LLaMA-3.2, Qwen2.5, and Mistral, with their sizes varying from 1B to 8B.
Experimental results demonstrate that our proposed bidirectional reconstruction training consistently improves performance across different models and scales. Notably, our method achieves new state-of-the-art results on the Massive Text Embeddings Benchmark (MTEB) \citep{muennighoff2023mteb} among models trained solely on publicly available data. Meanwhile, comprehensive ablation studies further validate the effectiveness of our proposed training objectives and the two-stage framework.


In summary, our key contributions are as follows:
\begin{itemize}
    \item We highlight a mismatch in the role of the \texttt{[EOS]} token between general language model pre-training and embedding tasks.
    \item We introduce a novel training stage consisting of two bidirectional generative reconstruction tasks, EBQ2D and EBD2Q, that encourage the model to inject semantic alignment into the \texttt{[EOS]} representation.
    \item Our approach consistently improves the quality of embeddings generated by LLMs, achieving new state-of-the-art results on MTEB.
\end{itemize}

\section{Related Works}
\paragraph{Text embeddings.} Text embeddings are vector representations of natural language text that encode its semantic content, which play a pivotal role in various natural language processing (NLP) tasks, such as information retrieval (IR), semantic similarity estimation, classification, and clustering \citep{fujiwara2023efficient,karpukhin2020dense}. As an example, the first-stage retrieval in an IR system leverages embedding similarity to retrieve relevant documents from a large-scale corpus. Apart from early attempts using latent semantic indexing \citep{deerwester1990indexing} and word-level representations \citep{mikolov2013efficient}, modern research on embedding task utilizes pre-trained language models, like BERT \citep{devlin2019bert}, RoBERTa \citep{liu2020roberta}, and T5 \citep{raffel2020exploring}, significantly outperforming traditional approaches. To further enhance the performance, advanced methods like E5 \citep{wang2022text} and BGE \citep{xiao2024c} employ a complex multi-stage training pipeline consisting of large-scale weakly supervised contrastive pre-training and multi-task fine-tuning. More recently, LLMs have become the new foundation for text embedding given their superior capability on semantic understanding \citep{brown2020language,chowdhery2023palm,touvron2023llama}. LLM2Vec \citep{behnamghader2024llm2vec} enables bidirectional attention and applies masked language modeling to transform decoder-only LLMs into text encoders. \citet{muennighoff2025generative} introduce an additional training objective to preserve generative capabilities, but still require bidirectional attention. Alternatively, Echo embeddings proposed by \citet{springer2025repetition} avoid architecture modifications and allow a unified model for embedding and generation. After repetition, the output embedding of the final token is adopted as the representation of the input text, consistent with other auto-regressive methods  \citep{li2025making, springer2025repetition, li2024llama2vec, wang2024improving}.

\paragraph{LLM-based retrieval.} LLM-based embedding models offer a strong backbone for retrieval systems, facilitating more precise modeling of complex relationships between queries and documents. 
Repllama \citep{ma2024fine} fine-tuned LLaMA-2 to function as both a retriever and a reranker, showcasing the potential of large language models in retrieval pipelines. Llama2Vec \citep{li2024llama2vec} further improved performance by introducing two pretext tasks, achieving significant gains on the BEIR \citep{thakur2021beir} benchmark. These methods similarly adopt the embedding of the final token as the overall representation of queries and documents. However, they overlook the discrepancy between the \texttt{[EOS]} token’s role in language model pre-training, where it functions merely as a sequence terminator, and its intended use as a semantic bridge in retrieval tasks, leaving the learning of more effective and query-document-aligned \texttt{[EOS]} embeddings an open challenge.


\paragraph{Auto-Reconstruction Methods.} Prior works such as SimLM \citep{wang-etal-2023-simlm}, LexMAE \citep{shen2022lexmae}, RetroMAE \citep{xiao2022retromae}, and Condenser \citep{gao2021condenser} have explored auto-reconstruction objectives to improve text embeddings for encoder-based models. To enrich the representations, these methods typically try to recover masked tokens or spans from intermediate encoder states. They rely on auxiliary decoders or reconstruction heads that are used during pre-training but discarded at inference, which often leads to complex multi-stage training pipelines.

\section{Method}

In this section, we present our two-stage training framework. The first stage is a novel training phase introduced in this work, which incorporates two bidirectional reconstruction tasks, detailed in Section~\ref{sec:stage_1}. The second stage employs contrastive learning to further refine the representations, as described in Section~\ref{sec:stage2}. An overview of the entire pipeline is illustrated in Figure~\ref{fig:pipeline}.

\subsection{Preliminary}
Language models (LMs) have been widely adopted as powerful embedding models in a variety of NLP tasks. A decoder-only language model \(\mathcal{M}\) typically consists of an input embedding layer \(\texttt{Embed}(\cdot)\), \(L\) stacked Transformer decoder blocks \(\texttt{Dec}_\ell(\cdot)\) with self-attention modules, and a linear output head \(\texttt{lm\_head}\) that projects the final hidden states to vocabulary logits for next-token prediction.

To obtain the embedding for an input sequence from LMs, we adopt a widely used approach that leverages the final hidden state. Specifically, given an input sequence \( X = \{x_1, x_2, \dots, x_n\} \), we append the special end-of-sequence token \texttt{\texttt{[EOS]}} to form the full input. The hidden state at the position of \texttt{\texttt{[EOS]}} extracted from the final decoder layer is taken as the sequence embedding, formally defined as:
\begin{equation}
\label{equ:embed}
    e_X = f(x_1, x_2, \dots, x_n, \texttt{\texttt{[EOS]}})[-1],
\end{equation}
where \(f(\cdot)\) denotes the composition of the embedding process and the forward pass through all $L$ stacked decoder layers, as computed below:
\begin{equation}
    f(\cdot) =  \texttt{Dec}_\ell(\texttt{Dec}_{\ell-1}(\dots \texttt{Dec}_1(\texttt{Embed}(\cdot)))).
\end{equation}

The objective of text retrieval is to find the top-k documents from a large-scale corpus that are most relevant to a given query. The semantic relevance between a query $Q$ and a candidate document $D$ is typically measured by the similarity between their embeddings, such as cosine similarity or the inner product, i.e., $⟨e_Q, e_D⟩$.


\subsection{Stage I: Bidirectional Reconstruction}
\label{sec:stage_1}

We introduce a novel training stage inserted between general LM pre-training and contrastive learning. During this stage, the model is supervised by two dual reconstruction objectives, namely EBQ2D (Embedding-Based Query-to-Document) and EBD2Q (Embedding-Based Document-to-Query), which guide the model to incorporate counterpart information into the output embedding. Accurate query-document pairs are used as the training corpus. The implementation details of the bidirectional reconstruction are illustrated in Algorithm~\ref{alg:anchor}.

\textbf{EBQ2D.}
Given a natural language query, the EBQ2D objective aims to encourage the model to generate the relevant document conditioned on the query embedding. During training, the model first computes the embedding \(e_Q\) from the query token sequence \( Q = \{q_1, \dots, q_n\} \), as described in Equation~\ref{equ:embed}. This embedding is then used as a prefix to condition the generation of document tokens \( D = \{d_1, \dots, d_m\} \) under the teacher forcing paradigm. The training objective minimizes the cross-entropy loss between the predicted and ground-truth tokens. This reconstruction of documents via query embedding requires the embedding to capture the explicit semantic meaning of the query while simultaneously integrating information indicative of the relevant document, which is essential for effectively retrieving relevant documents in subsequent tasks. Formally, the EBQ2D loss can be given by:
\begin{equation}
    \mathcal{L}_{\text{Q2D}} = - \sum_{t=1}^{m} \log P_{\Theta}(d_t \mid e_Q, d_{<t}).
\end{equation}
where $\Theta$ denotes the model parameters.

\begin{algorithm}[t]
\caption{Bidirectional Reconstruction}
\label{alg:anchor}
\begin{algorithmic}[1]
\Require Paired data $(Q, D)$; model $\mathcal{M} = \texttt{lm\_head} \circ \texttt{Dec} \circ \texttt{Embed}$

\State $e_Q \gets f(Q)$; $e_D \gets f(D)$

\Comment{Obtain embeddings as in Equation~\ref{equ:embed}}
\State $\mathbf{E}_Q \gets \texttt{Embed}(Q)$; $\mathbf{E}_D \gets \texttt{Embed}(D)$

\State $\hat{D} \gets \texttt{lm\_head}(\texttt{Dec}([\![e_Q, \mathbf{E}_D]\!]))$

\Comment{Decode $D$ from $e_Q$ via teacher forcing}
\State $\mathcal{L}_{\text{Q2D}} \gets \text{CE}(\hat{D}, D)$

\State $\hat{Q} \gets \texttt{lm\_head}(\texttt{Dec}([\![e_D, \mathbf{E}_Q]\!]))$

\Comment{Decode $Q$ from $e_D$ via teacher forcing}
\State $\mathcal{L}_{\text{D2Q}} \gets \text{CE}(\hat{Q}, Q)$

\State \Return $\mathcal{L}_{\text{StageI}} =  \alpha \mathcal{L}_{\text{Q2D}} + (1-\alpha) \mathcal{L}_{\text{D2Q}}$
\end{algorithmic}
\end{algorithm}

\textbf{EBD2Q.}
Complementary to EBQ2D, the EBD2Q objective guides the model to recover the underlying user intent from the given document. The model first encodes the document and uses the representation of the \texttt{\texttt{[EOS]}} token, denoted as \( e_D \), as the embedding that summarizes the document’s content. Conditioned on \( e_D \), the model generates the corresponding query auto-regressively, following a similar decoding process as in EBQ2D. This objective encourages the document embedding to capture high-level abstractions and latent intent signals necessary to reconstruct the query, enhancing the bidirectional alignment between queries and documents. Similarly, the EBD2Q loss is defined as:
\begin{equation}
\mathcal{L}_{\text{D2Q}} = - \sum_{t=1}^{n} \log P_{\Theta}(q_t \mid e_D, q_{<t}).
\end{equation}

Training Stage I integrates the two tasks within a multi-task learning framework, where the overall objective is formulated as a weighted sum of $\mathcal{L}_{Q2D}$ and $\mathcal{L}_{D2Q}$:
\begin{equation}
    \mathcal{L}_{\text{StageI}} =  \alpha \mathcal{L}_{\text{Q2D}} + (1-\alpha) \mathcal{L}_{\text{D2Q}}
\end{equation}
where \(\alpha \in [0, 1]\) is a hyperparameter that controls the relative importance of the two objectives. As our experiments show that the performance is not sensitive to the choice of $\alpha$, we fix it to 0.2, which yields slightly better results.

\subsection{Stage II: Contrastive Learning}
\label{sec:stage2}
After bidirectional reconstruction training, the model is fine-tuned on downstream tasks through contrastive learning. In line with prior work \citep{behnamghader2024llm2vec, springer2025repetition}, we use a replication of the public portion of the E5 dataset \citep{wang2024improving} as the training corpus for fine-tuning. The training process is guided by the widely used InfoNCE \citep{izacard2021unsupervised} loss function \(\mathcal{L}\):
\begin{equation}
\small
    \mathcal{L}=-\log \frac{\exp \left(\operatorname{sim}\left(Q, D^{+}\right)\right)}{\exp \left(\operatorname{sim}\left(Q, D^{+}\right)\right)+\sum_{j} \exp \left(\operatorname{sim}\left(Q, D_{j}^{-}\right)\right)}
\end{equation}

In this equation, \( D^{-}_j \) denotes the set of negatives, encompassing both in-batch and hard negatives. The matching score between a query \( Q \) and a document \( D \) is computed using a temperature-scaled cosine similarity function, defined as:
\begin{equation}
    \operatorname{sim}(Q, D) = \frac{1}{\tau}\cos(e_Q, e_D)
\end{equation}
where \(\tau\) is a temperature hyperparameter, which is fixed at 0.05 in our practice.

After our two-stage training, the embeddings generated by the model are enriched with semantic information and demonstrate improved alignment between queries and relevant documents. We refer to these enhanced representations as "Anchor Embeddings", which also denotes our proposed method.

\renewcommand{\arraystretch}{1.1}
\setlength{\tabcolsep}{4pt}
\begin{table*}[t]
\centering
\small
\begin{tabular}{lllllllll}
\toprule
\textbf{Categories $\rightarrow$} 
  & \multicolumn{1}{c}{\textbf{Retr.}} 
  & \multicolumn{1}{c}{\textbf{Rerank.}} 
  & \multicolumn{1}{c}{\textbf{Clust.}} 
  & \multicolumn{1}{c}{\textbf{PairClass.}} 
  & \multicolumn{1}{c}{\textbf{Class.}} 
  & \multicolumn{1}{c}{\textbf{STS}} 
  & \multicolumn{1}{c}{\textbf{Summ.}} 
  & \multicolumn{1}{c}{\textbf{Avg}} \\
\multicolumn{1}{l}{\# of datasets $\rightarrow$} 
  & \multicolumn{1}{c}{15} 
  & \multicolumn{1}{c}{4} 
  & \multicolumn{1}{c}{11} 
  & \multicolumn{1}{c}{3} 
  & \multicolumn{1}{c}{12} 
  & \multicolumn{1}{c}{10} 
  & \multicolumn{1}{c}{1} 
  & \multicolumn{1}{c}{56} \\
\midrule

\noalign{\vskip -0.2em} 
\multicolumn{9}{c}{\footnotesize \textbf{LLaMA-3.2-1B-Instruct}} \\
\noalign{\vskip -0.2em}
\midrule
Baseline & 50.06 & 54.94 & 44.38 & 82.71 & 72.17 & 81.27 & 29.94 & 60.99 \\
\rowcolor[gray]{0.9} Anchor (ours) & 52.60$_{\mathbf{+2.54}}$ & 56.65$_{\mathbf{+1.71}}$ & 44.75$_{+0.37}$ & 85.48$_{+2.77}$ & 72.47$_{+0.30}$ & 82.10$_{+0.83}$ & 30.87$_{+0.93}$ & 62.24$_{\mathbf{+1.25}}$ \\
\midrule

\noalign{\vskip -0.2em}
\multicolumn{9}{c}{\footnotesize \textbf{Qwen2.5-1.5B-Instruct}} \\
\noalign{\vskip -0.2em}
\midrule
Baseline & 51.66 & 54.86 & 43.03 & 84.41 & 72.97 & 82.02 & 31.89 & 61.51 \\
\rowcolor[gray]{0.9} Anchor (ours) & 53.62$_{\mathbf{+1.96}}$ & 57.63$_{\mathbf{+2.77}}$ & 43.19$_{+0.16}$ & 85.77$_{+1.36}$ & 74.51$_{+1.54}$ & 82.74$_{+0.72}$ & 31.61$_{-0.28}$ & 62.86$_{\mathbf{+1.35}}$ \\
\midrule

\noalign{\vskip -0.2em}
\multicolumn{9}{c}{\footnotesize \textbf{LLaMA-3.2-3B-Instruct}} \\
\noalign{\vskip -0.2em}
\midrule
Baseline & 51.72 & 56.13 & 43.40 & 86.11 & 74.67 & 82.73 & 30.98 & 62.33 \\
\rowcolor[gray]{0.9} Anchor (ours) & 53.66$_{\mathbf{+1.94}}$ & 57.77$_{\mathbf{+1.64}}$ & 45.48$_{+2.08}$ & 86.53$_{+0.42}$ & 75.59$_{+0.92}$ & 82.48$_{-0.25}$ & 30.81$_{-0.17}$ & 63.55$_{\mathbf{+1.22}}$ \\
\midrule

\noalign{\vskip -0.2em}
\multicolumn{9}{c}{\footnotesize \textbf{Mistral-7B}} \\
\noalign{\vskip -0.2em}
\midrule
Baseline & 54.92 & 57.98 & 44.97 & 86.04 & 75.51 & 83.14 & 30.64 & 63.87 \\
\rowcolor[gray]{0.9} Anchor (ours) & 56.87$_{\mathbf{+1.95}}$ & 60.56$_{\mathbf{+2.58}}$ & 45.73$_{+0.76}$ & 87.99$_{+1.95}$ & 75.95$_{+0.44}$ & 83.52$_{+0.38}$ & 30.28$_{-0.36}$ & 64.99$_{\mathbf{+1.12}}$ \\
\midrule

\noalign{\vskip -0.2em}
\multicolumn{9}{c}{\footnotesize \textbf{LLaMA-3.1-8B-Instruct}} \\
\noalign{\vskip -0.2em}
\midrule
Baseline & 55.36 & 58.92 & 46.64 & 86.80 & 74.80 & 83.10 & 29.67 & 64.06 \\
\rowcolor[gray]{0.9} Anchor (ours) & 57.09$_{\mathbf{+1.73}}$ & 61.38$_{\mathbf{+2.46}}$ & 46.03$_{-0.61}$ & 88.92$_{+2.12}$ & 76.17$_{+1.37}$ & 83.76$_{+0.66}$ & 30.13$_{+0.46}$ & 65.30$_{\mathbf{+1.24}}$ \\
\bottomrule
\end{tabular}
\caption{Performance on the MTEB benchmark. Baselines are trained only with regular contrastive learning (Stage II).}
\label{tab:anchor_vs_baselines}
\end{table*}
\renewcommand{\arraystretch}{1.1}
\setlength{\tabcolsep}{4pt}
\begin{table*}[h]
\centering
\small
\begin{tabular}{lcccccccc}
\toprule
\textbf{Categories $\rightarrow$} & \textbf{Retr.} & \textbf{Rerank.} & \textbf{Clust.} & \textbf{PairClass.} & \textbf{Class.} & \textbf{STS} & \textbf{Summ.} & \textbf{Avg} \\
\multicolumn{1}{l}{\# of datasets $\rightarrow$} 
& \multicolumn{1}{c}{15} 
& \multicolumn{1}{c}{4} 
& \multicolumn{1}{c}{11} 
& \multicolumn{1}{c}{3} 
& \multicolumn{1}{c}{12} 
& \multicolumn{1}{c}{10} 
& \multicolumn{1}{c}{1} 
& \multicolumn{1}{c}{56} \\
\midrule
\textbf{Previous work w/ public data only} \\
Instructor-xl & 49.26 & 57.29 & 44.74 & 86.62 & 73.12 & 83.06 & \textbf{32.32} & 61.79 \\
BGE$_{\text{large-en-v1.5}}$ & 54.29 & 60.03 & 46.08 & 87.12 & 75.97 & 83.11 & 31.61 & 64.23 \\
GritLM$_{\text{Mistral-7b-v1}}$ + public data & 53.10 & \underline{61.30} & \textbf{48.90} & 86.90 & \underline{77.00} & 82.80 & 29.40 & 64.70 \\
E5$_{\text{Mistral-7b-v1}}$ + public data & 52.78 & 60.38 & \underline{47.78} & \underline{88.47} & 76.80 & \underline{83.77} & \underline{31.90} & 64.56 \\
Echo$_{\text{Mistral-7b-v1}}$ & 55.52 & 58.14 & 46.32 & 87.34 & \textbf{77.43} & 82.56 & 30.73 & 64.68 \\
bge-en-icl$_{\text{Mistral-7b-v1}}$ + E5 data (zero-shot) & \textbf{59.59} & 56.85 & 42.61 & 87.87 &75.47 & 83.30 & 29.52 & 64.67 \\
LLM2Vec$_{\text{S-LLaMA-1.3B}}$ & 51.44 & 55.38 & 43.57 & 86.20 & 72.21 & 83.58 & 30.01 & 61.85 \\
LLM2Vec$_{\text{Mistral-7B}}$ & 55.99 & 58.42 & 45.54 & 87.99 & 76.63 & \textbf{84.09} & 29.96 & 64.80 \\
LLM2Vec$_{\text{Meta-LLaMA-3-8B}}$ & 56.63 & 59.68 & 46.45 & 87.80 & 75.92 & 83.58 & 30.94 & \underline{65.01} \\
\midrule
\textbf{Anchor$_{\text{LLaMA-3.1-8B-Instruct}}$} & \underline{57.09} & \textbf{61.38} & 46.03 & \textbf{88.92} & 76.17 & 83.76 & 30.13 & \textbf{65.30} \\
\bottomrule
\end{tabular}
\caption{Performance comparison on the MTEB benchmark with other advanced models. The best results for each subtask are highlighted in bold, and the second-best results are underlined.}
\label{tab:anchor_vs_sota}
\end{table*}

\section{Experiment}
\subsection{Basic Settings}

\paragraph{Language Models.} We apply our two-stage training framework to five decoder-only LLMs ranging from 1B to 8B parameters: Meta-LLaMA-3.2-1B-Instruct, Qwen2.5-1.5B-Instruct, Meta-LLaMA-3.2-3B-Instruct, Mistral-7B, and Meta-LLaMA-3.1-8B-Instruct.

\paragraph{Training Datasets and Setup.} The public portion of the E5 dataset \citep{wang2024improving}, curated by \citet{springer2025repetition}, serves as the training corpus for both stages. It comprises roughly 1.5 million samples, with further details on its construction provided in \ref{sec:e5}. 
We train the model with full-parameter tuning for 2000 steps in Stage I, and fine-tune it using LoRA for 1000 steps as done in LLM2Vec \citep{behnamghader2024llm2vec} in Stage II. We provide other hyper-parameters in Appendix~\ref{sec:train_detail}.

\paragraph{Benchmark.} We evaluate our method on the Massive Text Embedding Benchmark (MTEB) \cite{muennighoff2023mteb}, a collection of 56 datasets covering seven types of embedding tasks: classification, clustering, pairwise classification, re-ranking, retrieval, sentence similarity (STS), and summarization. A comprehensive description of MTEB is provided in Appendix~\ref{sec:mteb}. Since MTEB is massive and requires multiple days to evaluate, we conduct ablations on a 15-task subset as adopted in LLM2Vec, with details provided in Appendix~\ref{sec:mteb_subset}.

\paragraph{Baselines.} Since different models are often trained on diverse datasets, and many do not disclose the specific data used, for a fairer comparison and to more accurately assess the impact of our proposed training strategy, we conduct evaluations comparing against models trained solely on the publicly available data under zero-shot settings. The compared methods include earlier encoder-based models, such as Instructor-xl \citep{su2023one} (1.5B) and BGE-large-en-v1.5 \citep{xiao2024c} (335M). In addition, we compare against recent state-of-the-art approaches, including GritLM \citep{muennighoff2024generative}, E5 \citep{wang2024improving}, bge-en-icl \citep{li2025making}, and the fine-tuned Echo embedding \citep{springer2025repetition}, all of which are built upon Mistral-7B. For LLM2Vec \citep{behnamghader2024llm2vec}, we compare against its \texttt{Bi + MNTP} variants built on S-LLaMA-1.3B, Mistral-7B, and Meta-LLaMA-3-8B, which achieve the best performance after supervised fine-tuning.

\subsection{Main Results}

Table~\ref{tab:anchor_vs_baselines} presents the performance of our method (marked as Anchor) compared to baselines trained solely with contrastive learning, without incorporating the proposed bidirectional reconstruction stage. All evaluations are conducted on the MTEB benchmark, with improvements over the baseline indicated as subscripts. Table~\ref{tab:anchor_vs_sota} further compares our method against other state-of-the-art models on MTEB. Based on these tables, we analyze the results from the following three perspectives.

Firstly, the bidirectional reconstruction training consistently improves performance across different models and scales. For instance, compared to the baseline LLaMA-3.2-1B-Instruct model which is only fine-tuned with contrastive learning, Anchor$_{\text{LLaMA-3.2-1B-Instruct}}$ achieves an average score of 62.24\%, with an absolute improvement of 1.25\%. Especially, it shows substantial gains of 2.54\% and 1.71\% on the retrieval and re-ranking tasks, respectively. Similarly, Anchor$_{\text{Qwen2.5-1.5B-Instruct}}$ and Anchor$_{\text{Mistral-7B}}$ also demonstrate clear improvements, further validating the robustness of our method across different model families. Notably, for larger models such as Mistral-7B, LLaMA-3.2-3B-Instruct, and LLaMA-3.1-8B-Instruct, our method continues to yield consistent gains, with average improvements of +1.12\%, +1.22\%, and +1.24\%, respectively. This indicates that our bidirectional reconstruction training maintains its effectiveness as model size increases, which is essential to achieve superior performance with stronger models.

Secondly, our method establishes a new state-of-the-art performance on the MTEB. At the 1B scale, Anchor$_{\text{LLaMA-3.2-1B-Instruct}}$ and Anchor$_{\text{Qwen2.5-1.5B-Instruct}}$ achieve average scores of 62.24\% and 62.86\%, respectively, outperforming LLM2Vec$_{\text{S-LLaMA-1.3B}}$ at 61.85\%. Likewise, Anchor$_{\text{Mistral-7B}}$ attains 64.99\%, surpassing not only its contrastive baseline but also other competitive Mistral-7B-based models such as LLM2Vec (64.80\%), GritLM (64.70\%), E5 (64.56\%), and Echo (64.68\%), thereby confirming the effectiveness of our method under the same backbone architecture. At the larger scale, Anchor$_{\text{LLaMA-3.1-8B-Instruct}}$ reaches 65.30\%, exceeding LLM2Vec with the same 8B scale at 65.01\%, and establishing a new state-of-the-art performance. These results validate that our approach promotes the learning of richer semantic representations, yielding higher-quality embeddings and enhanced downstream task performance.

Thirdly, the bidirectional reconstruction tasks generally lead to performance improvements across most sub-tasks. The EBQ2D and EBD2Q objectives were specifically designed to capture implicit semantic relationships between queries and documents, primarily targeting retrieval and re-ranking tasks. Unexpectedly, we find that the reconstruction training maintains the model performance on other tasks and, in some cases, even delivers slight improvements. For instance, with LLaMA-3.2-1B-Instruct, our approach yields performance gains across all task categories, with the most notable improvements observed in retrieval and re-ranking as expected. This suggests that by learning to reconstruct the counterpart text from the \texttt{[EOS]} embedding, the model is encouraged to encode a more compact and semantically rich representation of the input, while better capturing relational information. For instance, for pair classification task, it becomes more effective at representing the relationship between sentence pairs, aiding in tasks such as assessing semantic similarity.

\begin{figure*}[t]
    \centering
    \subfigure[LLaMA-3.2-1B-Instruct]{
        \includegraphics[width=0.31\textwidth]{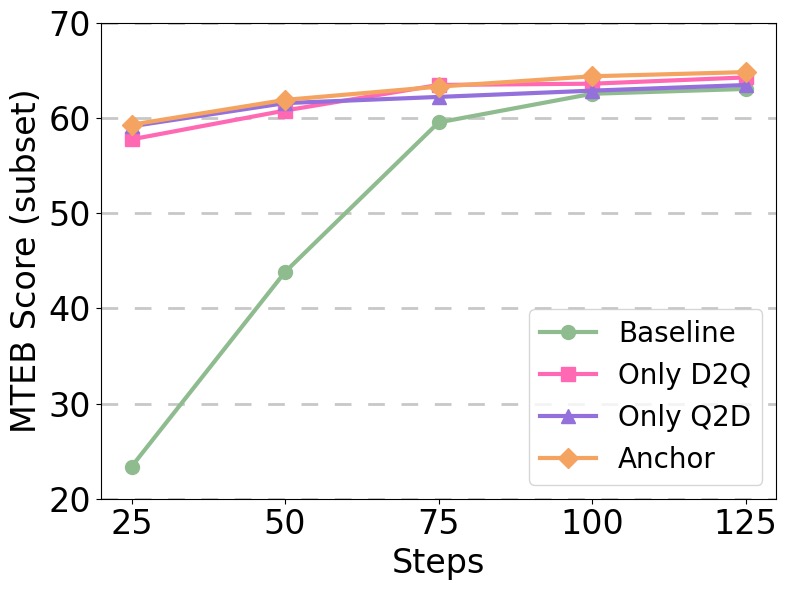}
    }
    \subfigure[LLaMA-3.2-3B-Instruct]{
        \includegraphics[width=0.31\textwidth]{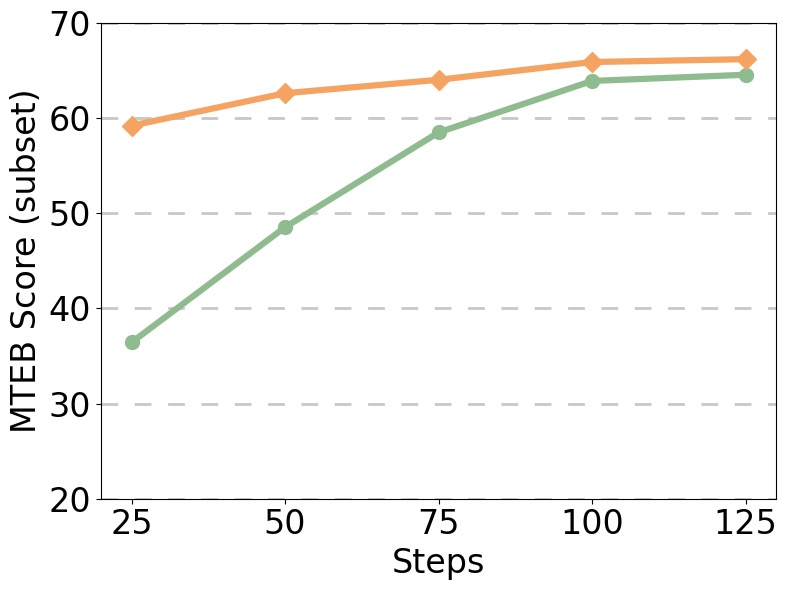}
    }
    \subfigure[LLaMA-3.1-8B-Instruct]{
        \includegraphics[width=0.31\textwidth]{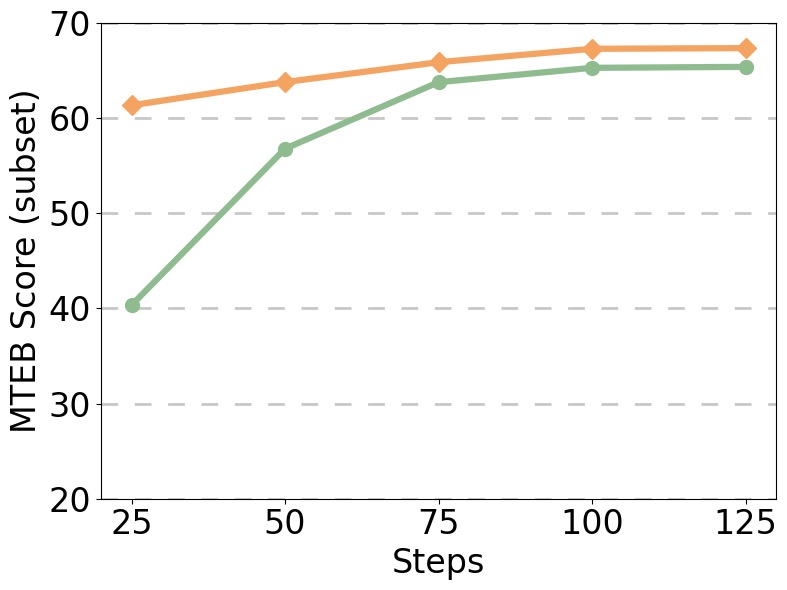}
    }
    \caption{Results on the 15 task subset of MTEB during the first 125 training steps for LLaMA-3.2-1B-Instruct, LLaMA-3.2-3B-Instruct, and LLaMA-3.1-8B-Instruct.}
    \label{fig:convergence}
\end{figure*}

\subsection{Ablation Study}

\paragraph{Ablation on Reconstruction Tasks. } 
We evaluate the effectiveness of the two reconstruction tasks proposed in Stage I, using the LLaMA-3.2-1B-Instruct model. The results are summarized in Table \ref{tab:ablation}. The baseline refers to models trained exclusively with contrastive learning (i.e., without the bidirectional reconstruction stage), while "OnlyD2Q" and "OnlyQ2D" represent models trained with only one of the two reconstruction objectives during Stage I. Compared to the baseline, both the OnlyD2Q and OnlyQ2D variants yield consistent improvements across most tasks. In particular, the OnlyQ2D variant achieves higher scores. This indicates that the challenge of generating comprehensive documents from brief queries promotes the model to produce more expressive informative text embeddings. Moreover, our proposed method, which integrates both objectives, achieves the best overall performance with the highest average score of 62.24. 

To investigate the impact of the hyperparameter $\alpha$, we conduct experiments on the MTEB subset with $\alpha$ set to 0.2, 0.5, and 0.8. As shown in Table~\ref{tab:alpha-ablation}, the model exhibits relatively stable performance across these values, with our selected value $\alpha = 0.2$ yielding the best results.
\renewcommand{\arraystretch}{1.1}
\begin{table}[t]
\centering
\small
\begin{tabular}{lcccc}
\toprule
\textbf{Models $\rightarrow$} & \textbf{Baseline} & \textbf{OnlyD2Q} & \textbf{OnlyQ2D} & \textbf{Anchor} \\
$\bm{\alpha}\rightarrow$        & -  & 0 & 1     & 0.2      \\
\midrule
\textbf{Retr.}       & 50.06 & 51.54 & 52.05 & \textbf{52.60} \\
\textbf{Rerank.}     & 54.94 & 56.30 & 56.62 & \textbf{56.65} \\
\textbf{Clust.}      & 44.38 & 43.59 & 44.04 & \textbf{44.75} \\
\textbf{PairClass.}  & 82.71 & 85.12 & \textbf{85.50} & 85.48 \\
\textbf{Class.}      & 72.17 & 72.26 & 71.83 & \textbf{72.47} \\
\textbf{STS}         & 81.27 & 80.70 & 81.04 & \textbf{82.10} \\
\textbf{Summ.}       & 29.94 &\textbf{ 30.91} & 30.24 & 30.87 \\
\midrule
\textbf{Avg}         & 60.99 & 61.38 & 61.62 & \textbf{62.24} \\
\bottomrule
\end{tabular}
\caption{Ablation study of training objectives in Stage I over the full MTEB benchmark.}
\label{tab:ablation}
\end{table}

\renewcommand{\arraystretch}{1.1}
\setlength{\tabcolsep}{4pt}
\begin{table}[h]
\centering
\begin{tabular}{c|c}
\toprule
\textbf{$\bm{\alpha}$} & \textbf{Average Score} \\
\hline
0.2 & \textbf{65.19} \\
0.5 & 64.88 \\
0.8 & 65.12 \\
\bottomrule
\end{tabular}
\caption{Performance under different $\alpha$ values.}
\label{tab:alpha-ablation}
\end{table}

\paragraph{Early-stage During Fine-tuning.} We save checkpoints every 25 steps and evaluate on the 15-task MTEB subset to assess early-stage fine-tuning performance. As shown in Figure~\ref{fig:convergence}, models trained with our bidirectional reconstruction tasks consistently demonstrate stronger performance in the early stages of fine-tuning in Stage II across all model sizes. The models trained after Stage I are nearly converged at the very beginning of fine-tuning, as further evidenced by the loss curves in Figure~\ref{fig:train_loss} provided in the Appendix. This suggests that our training with bidirectional reconstruction tasks is sufficiently powerful to endow the model with high-quality textual representations, reducing the reliance on subsequent contrastive learning.

\begin{figure}
\centering
\includegraphics[width=0.4\textwidth]{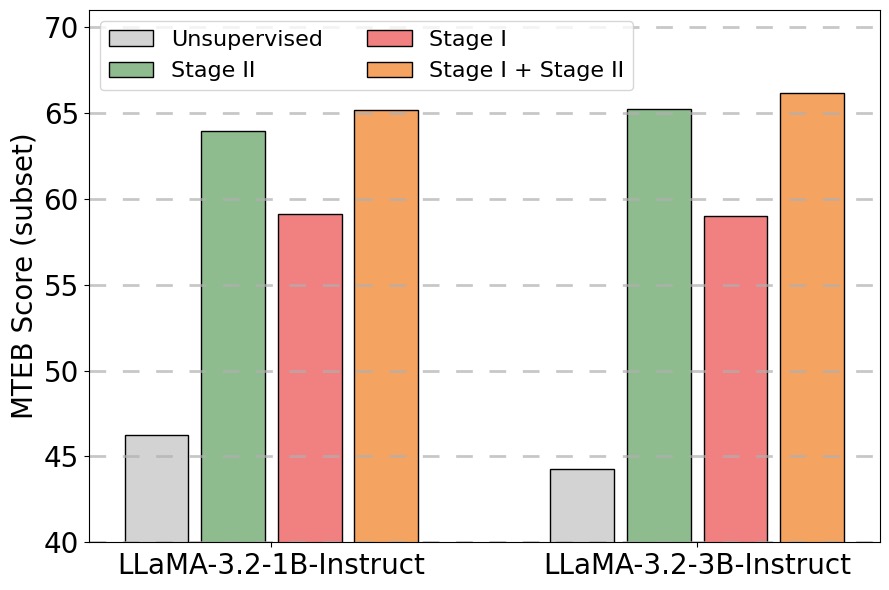}
\caption{Impact of the two training stages on performance.}
\label{fig:stage}
\end{figure}

\paragraph{Training stage.} We conduct an ablation study to assess the contribution of each training stage, with the results plotted in Figure~\ref{fig:stage}. The unsupervised baseline yields relatively low performance, suggesting that without task-specific supervision, the \texttt{[EOS]} output embeddings are insufficient for downstream tasks. Applying only Stage I with bidirectional reconstruction tasks results in a notable improvement, demonstrating that the model begins to learn semantic alignment through this training process. Incorporating Stage II leads to a further improvement, highlighting the indispensable role of contrastive learning in producing high-quality embeddings. Our two-stage framework significantly outperforms both single-stage variants: Stage I helps the model encode implicit semantic alignment between relevant texts, while Stage II further refines the representation space by bringing similar instances closer and pushing dissimilar ones apart. 

\renewcommand{\arraystretch}{1.1}
\begin{table}[t]
\centering
\label{tab:extra-training}
\begin{tabular}{lll}
\toprule
\textbf{Training Configuration} & \textbf{Steps} & \textbf{Score} \\
\midrule
Contrastive only (Stage II) & 1,000 & 63.95 \\
Contrastive only (Stage II) & 3,000 & 63.74 \\
Full method (Stage I + II) & 2,000 + 1,000 & \textbf{65.19} \\
\bottomrule
\end{tabular}
\caption{Comparison of training strategies on the MTEB subset using LLaMA-3.2-1B-Instruct.}
\label{tab:not_step}
\end{table}

\paragraph{Effectiveness beyond extra training.}
To rule out the possibility that the performance gains of our method stem merely from additional training steps, we evaluated a checkpoint trained solely on Stage II for 3,000 steps using the LLaMA-3.2-1B-Instruct model on an MTEB subset. The result was 63.74 as shown in Table~\ref{tab:not_step}, which is notably lower than the 65.19 achieved by our full method (Stage I: 2,000 steps + Stage II: 1,000 steps). Interestingly, the 3,000-step baseline even underperforms the 1,000-step version (63.95), which supports the rationale behind following LLM2Vec in reporting results after 1,000 steps of contrastive learning. These findings indicate that the observed improvement arises from the effectiveness of our bidirectional reconstruction training, rather than simply from longer training.

\section{Conclusion}

In this paper, we propose a two-stage training procedure for applying LLMs on text embedding tasks. Our method is designed to address the mismatch between the role of the \texttt{[EOS]} token in language model pre-training and downstream embedding tasks. We introduce two bidirectional reconstruction objectives, EBQ2D and EBD2Q, which treat the \texttt{[EOS]} output embedding as an anchor to aggregate semantic information from either queries or relevant documents to reconstruct their counterparts. This stage encourages the model to encode semantic alignment directly into the \texttt{[EOS]} representation. Our method achieves state-of-the-art performance on MTEB among models trained on the same publicly available datasets under zero-shot settings. We hope that our method offers valuable insight to advance the development and application of embedding models.

\section*{Limitations}
While our method has effectively enhanced the performance of LLMs as text embedders, the current work can still be improved in the following
ways. First, the query-document bidirectional reconstruction tasks in Stage I are primarily designed to benefit retrieval and re-ranking. For broader embedding tasks such as classification or clustering, more tailored objectives may yield further improvements. Secondly, although the first-stage training accelerates convergence in the second-stage fine-tuning and can reduce its computational cost, the overall two-stage training framework still introduces additional overhead. Future work should investigate how to improve training efficiency in order to support scaling to larger models. Thirdly, the current model is primarily developed for English-centric scenarios, and expanding its applicability to other languages remains an important future direction.

\section*{Ethical Considerations}
Although our approach improves text embedding quality, it does not eliminate potential ethical risks associated with large language models. First, our approach may still inherit biases present in the training data, which can be amplified during the representation learning process, especially in domain-specific or underrepresented contexts. These biases can negatively influence downstream applications, such as search or recommendation systems. Additionally, like other LLM-based methods, our model may generate misleading or hallucinated outputs when used in generative scenarios, posing risks in high-stakes applications. We encourage responsible use and further evaluation of the model's behavior, particularly in sensitive domains.

\section*{Acknowledgments}
This research work has been sponsored by National Key Research and Development Program of China (No. 2023ZD0121402), Ant Group Security and Risk Management Fund, and National Natural Science Foundation of China (NSFC) grant (No. 62576211).

\bibliography{custom}

\begin{thebibliography}{47}
\providecommand{\natexlab}[1]{#1}

\bibitem[{BehnamGhader et~al.(2024)BehnamGhader, Adlakha, Mosbach, Bahdanau, Chapados, and Reddy}]{behnamghader2024llm2vec}
Parishad BehnamGhader, Vaibhav Adlakha, Marius Mosbach, Dzmitry Bahdanau, Nicolas Chapados, and Siva Reddy. 2024.
\newblock Llm2vec: Large language models are secretly powerful text encoders.
\newblock \emph{arXiv preprint arXiv:2404.05961}.

\bibitem[{Brown et~al.(2020)Brown, Mann, Ryder, Subbiah, Kaplan, Dhariwal, Neelakantan, Shyam, Sastry, Askell et~al.}]{brown2020language}
Tom Brown, Benjamin Mann, Nick Ryder, Melanie Subbiah, Jared~D Kaplan, Prafulla Dhariwal, Arvind Neelakantan, Pranav Shyam, Girish Sastry, Amanda Askell, and 1 others. 2020.
\newblock Language models are few-shot learners.
\newblock \emph{Advances in neural information processing systems}, 33:1877--1901.

\bibitem[{Chowdhery et~al.(2023)Chowdhery, Narang, Devlin, Bosma, Mishra, Roberts, Barham, Chung, Sutton, Gehrmann et~al.}]{chowdhery2023palm}
Aakanksha Chowdhery, Sharan Narang, Jacob Devlin, Maarten Bosma, Gaurav Mishra, Adam Roberts, Paul Barham, Hyung~Won Chung, Charles Sutton, Sebastian Gehrmann, and 1 others. 2023.
\newblock Palm: Scaling language modeling with pathways.
\newblock \emph{Journal of Machine Learning Research}, 24(240):1--113.

\bibitem[{DataCanary et~al.(2017)DataCanary, hilfialkaff, Jiang, Risdal, Dandekar, and tomtung}]{quora-question-pairs}
DataCanary, hilfialkaff, Lili Jiang, Meg Risdal, Nikhil Dandekar, and tomtung. 2017.
\newblock Quora question pairs.
\newblock \url{https://kaggle.com/competitions/quora-question-pairs}.
\newblock Kaggle.

\bibitem[{Deerwester et~al.(1990)Deerwester, Dumais, Furnas, Landauer, and Harshman}]{deerwester1990indexing}
Scott Deerwester, Susan~T Dumais, George~W Furnas, Thomas~K Landauer, and Richard Harshman. 1990.
\newblock Indexing by latent semantic analysis.
\newblock \emph{Journal of the American society for information science}, 41(6):391--407.

\bibitem[{Devlin et~al.(2019)Devlin, Chang, Lee, and Toutanova}]{devlin2019bert}
Jacob Devlin, Ming-Wei Chang, Kenton Lee, and Kristina Toutanova. 2019.
\newblock Bert: Pre-training of deep bidirectional transformers for language understanding.
\newblock In \emph{Proceedings of the 2019 conference of the North American chapter of the association for computational linguistics: human language technologies, volume 1 (long and short papers)}, pages 4171--4186.

\bibitem[{Fan et~al.(2019)Fan, Jernite, Perez, Grangier, Weston, and Auli}]{fan2019eli5}
Angela Fan, Yacine Jernite, Ethan Perez, David Grangier, Jason Weston, and Michael Auli. 2019.
\newblock Eli5: Long form question answering.
\newblock In \emph{Proceedings of the 57th Annual Meeting of the Association for Computational Linguistics}, pages 3558--3567.

\bibitem[{Fujiwara et~al.(2023)Fujiwara, Ida, Kumagai, Nakano, Kimura, and Ueda}]{fujiwara2023efficient}
Yasuhiro Fujiwara, Yasutoshi Ida, Atsutoshi Kumagai, Masahiro Nakano, Akisato Kimura, and Naonori Ueda. 2023.
\newblock Efficient network representation learning via cluster similarity.
\newblock \emph{Data Science and Engineering}, 8(3):279--291.

\bibitem[{Gao and Callan(2021)}]{gao2021condenser}
Luyu Gao and Jamie Callan. 2021.
\newblock Condenser: a pre-training architecture for dense retrieval.
\newblock In \emph{Proceedings of the 2021 Conference on Empirical Methods in Natural Language Processing}.

\bibitem[{Gao et~al.(2021)Gao, Yao, and Chen}]{gao2021simcse}
Tianyu Gao, Xingcheng Yao, and Danqi Chen. 2021.
\newblock Simcse: Simple contrastive learning of sentence embeddings.
\newblock In \emph{Proceedings of the 2021 Conference on Empirical Methods in Natural Language Processing}, page 6894. Association for Computational Linguistics.

\bibitem[{G{\"u}nther et~al.(2023{\natexlab{a}})G{\"u}nther, Milliken, Geuter, Mastrapas, Wang, and Xiao}]{gunther2023jina}
Michael G{\"u}nther, Louis Milliken, Jonathan Geuter, Georgios Mastrapas, Bo~Wang, and Han Xiao. 2023{\natexlab{a}}.
\newblock Jina embeddings: A novel set of high-performance sentence embedding models.
\newblock \emph{arXiv preprint arXiv:2307.11224}.

\bibitem[{G{\"u}nther et~al.(2023{\natexlab{b}})G{\"u}nther, Ong, Mohr, Abdessalem, Abel, Akram, Guzman, Mastrapas, Sturua, Wang et~al.}]{gunther2023jina2}
Michael G{\"u}nther, Jackmin Ong, Isabelle Mohr, Alaeddine Abdessalem, Tanguy Abel, Mohammad~Kalim Akram, Susana Guzman, Georgios Mastrapas, Saba Sturua, Bo~Wang, and 1 others. 2023{\natexlab{b}}.
\newblock Jina embeddings 2: 8192-token general-purpose text embeddings for long documents.
\newblock \emph{arXiv preprint arXiv:2310.19923}.

\bibitem[{He et~al.(2018)He, Liu, Liu, Lyu, Zhao, Xiao, Liu, Wang, Wu, She et~al.}]{he2018dureader}
Wei He, Kai Liu, Jing Liu, Yajuan Lyu, Shiqi Zhao, Xinyan Xiao, Yuan Liu, Yizhong Wang, Hua Wu, Qiaoqiao She, and 1 others. 2018.
\newblock Dureader: a chinese machine reading comprehension dataset from real-world applications.
\newblock In \emph{Proceedings of the Workshop on Machine Reading for Question Answering}. Association for Computational Linguistics.

\bibitem[{Izacard et~al.(2021)Izacard, Caron, Hosseini, Riedel, Bojanowski, Joulin, and Grave}]{izacard2021unsupervised}
Gautier Izacard, Mathilde Caron, Lucas Hosseini, Sebastian Riedel, Piotr Bojanowski, Armand Joulin, and Edouard Grave. 2021.
\newblock Unsupervised dense information retrieval with contrastive learning.
\newblock \emph{arXiv preprint arXiv:2112.09118}.

\bibitem[{Joshi et~al.(2017)Joshi, Choi, Weld, and Zettlemoyer}]{joshi2017triviaqa}
Mandar Joshi, Eunsol Choi, Daniel Weld, and Luke Zettlemoyer. 2017.
\newblock Triviaqa: A large scale distantly supervised challenge dataset for reading comprehension.
\newblock In \emph{Proceedings of the 55th Annual Meeting of the Association for Computational Linguistics (Volume 1: Long Papers)}. Association for Computational Linguistics.

\bibitem[{Karpukhin et~al.(2020)Karpukhin, Oguz, Min, Lewis, Wu, Edunov, Chen, and Yih}]{karpukhin2020dense}
Vladimir Karpukhin, Barlas Oguz, Sewon Min, Patrick~SH Lewis, Ledell Wu, Sergey Edunov, Danqi Chen, and Wen-tau Yih. 2020.
\newblock Dense passage retrieval for open-domain question answering.
\newblock In \emph{EMNLP (1)}, pages 6769--6781.

\bibitem[{Lee et~al.(2024)Lee, Roy, Xu, Raiman, Shoeybi, Catanzaro, and Ping}]{lee2024nv}
Chankyu Lee, Rajarshi Roy, Mengyao Xu, Jonathan Raiman, Mohammad Shoeybi, Bryan Catanzaro, and Wei Ping. 2024.
\newblock Nv-embed: Improved techniques for training llms as generalist embedding models.
\newblock \emph{arXiv preprint arXiv:2405.17428}.

\bibitem[{Lee et~al.(2025)Lee, Chen, Dua, Cer, Shanbhogue, Naim, {\'A}brego, Li, Chen, Vera et~al.}]{lee2025gemini}
Jinhyuk Lee, Feiyang Chen, Sahil Dua, Daniel Cer, Madhuri Shanbhogue, Iftekhar Naim, Gustavo~Hern{\'a}ndez {\'A}brego, Zhe Li, Kaifeng Chen, Henrique~Schechter Vera, and 1 others. 2025.
\newblock Gemini embedding: Generalizable embeddings from gemini.
\newblock \emph{arXiv preprint arXiv:2503.07891}.

\bibitem[{Lewis et~al.(2020)Lewis, Perez, Piktus, Petroni, Karpukhin, Goyal, K{\"u}ttler, Lewis, Yih, Rockt{\"a}schel et~al.}]{lewis2020retrieval}
Patrick Lewis, Ethan Perez, Aleksandra Piktus, Fabio Petroni, Vladimir Karpukhin, Naman Goyal, Heinrich K{\"u}ttler, Mike Lewis, Wen-tau Yih, Tim Rockt{\"a}schel, and 1 others. 2020.
\newblock Retrieval-augmented generation for knowledge-intensive nlp tasks.
\newblock \emph{Advances in neural information processing systems}, 33:9459--9474.

\bibitem[{Li et~al.(2024{\natexlab{a}})Li, Liu, Xiao, Shao, and Lian}]{li2024llama2vec}
Chaofan Li, Zheng Liu, Shitao Xiao, Yingxia Shao, and Defu Lian. 2024{\natexlab{a}}.
\newblock Llama2vec: Unsupervised adaptation of large language models for dense retrieval.
\newblock In \emph{Proceedings of the 62nd Annual Meeting of the Association for Computational Linguistics (Volume 1: Long Papers)}, pages 3490--3500.

\bibitem[{Li et~al.(2025)Li, Qin, Xiao, Chen, Luo, Lian, Shao, and Liu}]{li2025making}
Chaofan Li, Minghao Qin, Shitao Xiao, Jianlyu Chen, Kun Luo, Defu Lian, Yingxia Shao, and Zheng Liu. 2025.
\newblock \href {https://openreview.net/forum?id=wfLuiDjQ0u} {Making text embedders few-shot learners}.
\newblock In \emph{The Thirteenth International Conference on Learning Representations}.

\bibitem[{Li et~al.(2024{\natexlab{b}})Li, Tang, Chen, and Chen}]{li2024conan}
Shiyu Li, Yang Tang, Shizhe Chen, and Xi~Chen. 2024{\natexlab{b}}.
\newblock Conan-embedding: General text embedding with more and better negative samples.
\newblock \emph{arXiv preprint arXiv:2408.15710}.

\bibitem[{Li et~al.(2023)Li, Zhang, Zhang, Long, Xie, and Zhang}]{li2023towards}
Zehan Li, Xin Zhang, Yanzhao Zhang, Dingkun Long, Pengjun Xie, and Meishan Zhang. 2023.
\newblock Towards general text embeddings with multi-stage contrastive learning.
\newblock \emph{arXiv preprint arXiv:2308.03281}.

\bibitem[{Liu et~al.(2020)Liu, Ott, Goyal, Du, Joshi, Chen, Levy, Lewis, Zettlemoyer, and Stoyanov}]{liu2020roberta}
Yinhan Liu, Myle Ott, Naman Goyal, Jingfei Du, Mandar Joshi, Danqi Chen, Omer Levy, Mike Lewis, Luke Zettlemoyer, and Veselin Stoyanov. 2020.
\newblock \href {https://openreview.net/forum?id=SyxS0T4tvS} {Ro{\{}bert{\}}a: A robustly optimized {\{}bert{\}} pretraining approach}.

\bibitem[{Ma et~al.(2024)Ma, Wang, Yang, Wei, and Lin}]{ma2024fine}
Xueguang Ma, Liang Wang, Nan Yang, Furu Wei, and Jimmy Lin. 2024.
\newblock Fine-tuning llama for multi-stage text retrieval.
\newblock In \emph{Proceedings of the 47th International ACM SIGIR Conference on Research and Development in Information Retrieval}, pages 2421--2425.

\bibitem[{Mikolov et~al.(2013)Mikolov, Chen, Corrado, and Dean}]{mikolov2013efficient}
Tomas Mikolov, Kai Chen, Greg Corrado, and Jeffrey Dean. 2013.
\newblock Efficient estimation of word representations in vector space.
\newblock \emph{arXiv preprint arXiv:1301.3781}.

\bibitem[{Muennighoff et~al.(2024)Muennighoff, SU, Wang, Yang, Wei, Yu, Singh, and Kiela}]{muennighoff2024generative}
Niklas Muennighoff, Hongjin SU, Liang Wang, Nan Yang, Furu Wei, Tao Yu, Amanpreet Singh, and Douwe Kiela. 2024.
\newblock \href {https://openreview.net/forum?id=8cQrRO9iFe} {Generative representational instruction tuning}.
\newblock In \emph{ICLR 2024 Workshop: How Far Are We From AGI}.

\bibitem[{Muennighoff et~al.(2025)Muennighoff, SU, Wang, Yang, Wei, Yu, Singh, and Kiela}]{muennighoff2025generative}
Niklas Muennighoff, Hongjin SU, Liang Wang, Nan Yang, Furu Wei, Tao Yu, Amanpreet Singh, and Douwe Kiela. 2025.
\newblock \href {https://openreview.net/forum?id=BC4lIvfSzv} {Generative representational instruction tuning}.
\newblock In \emph{The Thirteenth International Conference on Learning Representations}.

\bibitem[{Muennighoff et~al.(2023)Muennighoff, Tazi, Magne, and Reimers}]{muennighoff2023mteb}
Niklas Muennighoff, Nouamane Tazi, Loic Magne, and Nils Reimers. 2023.
\newblock Mteb: Massive text embedding benchmark.
\newblock In \emph{Proceedings of the 17th Conference of the European Chapter of the Association for Computational Linguistics}, pages 2014--2037.

\bibitem[{Nguyen et~al.(2016)Nguyen, Rosenberg, Song, Gao, Tiwary, Majumder, and Deng}]{nguyen2016ms}
Tri Nguyen, Mir Rosenberg, Xia Song, Jianfeng Gao, Saurabh Tiwary, Rangan Majumder, and Li~Deng. 2016.
\newblock Ms marco: A human-generated machine reading comprehension dataset.

\bibitem[{Raffel et~al.(2020)Raffel, Shazeer, Roberts, Lee, Narang, Matena, Zhou, Li, and Liu}]{raffel2020exploring}
Colin Raffel, Noam Shazeer, Adam Roberts, Katherine Lee, Sharan Narang, Michael Matena, Yanqi Zhou, Wei Li, and Peter~J Liu. 2020.
\newblock Exploring the limits of transfer learning with a unified text-to-text transformer.
\newblock \emph{Journal of machine learning research}, 21(140):1--67.

\bibitem[{Rajpurkar et~al.(2016)Rajpurkar, Zhang, Lopyrev, and Liang}]{rajpurkar2016squad}
Pranav Rajpurkar, Jian Zhang, Konstantin Lopyrev, and Percy Liang. 2016.
\newblock Squad: 100,000+ questions for machine comprehension of text.
\newblock In \emph{Proceedings of the 2016 Conference on Empirical Methods in Natural Language Processing}, pages 2383--2392.

\bibitem[{Shen et~al.(2022)Shen, Geng, Tao, Xu, Huang, Jiao, Yang, and Jiang}]{shen2022lexmae}
Tao Shen, Xiubo Geng, Chongyang Tao, Can Xu, Xiaolong Huang, Binxing Jiao, Linjun Yang, and Daxin Jiang. 2022.
\newblock Lexmae: Lexicon-bottlenecked pretraining for large-scale retrieval.
\newblock \emph{arXiv preprint arXiv:2208.14754}.

\bibitem[{Springer et~al.(2025)Springer, Kotha, Fried, Neubig, and Raghunathan}]{springer2025repetition}
Jacob~Mitchell Springer, Suhas Kotha, Daniel Fried, Graham Neubig, and Aditi Raghunathan. 2025.
\newblock \href {https://openreview.net/forum?id=Ahlrf2HGJR} {Repetition improves language model embeddings}.
\newblock In \emph{The Thirteenth International Conference on Learning Representations}.

\bibitem[{Su et~al.(2023)Su, Shi, Kasai, Wang, Hu, Ostendorf, Yih, Smith, Zettlemoyer, and Yu}]{su2023one}
Hongjin Su, Weijia Shi, Jungo Kasai, Yizhong Wang, Yushi Hu, Mari Ostendorf, Wen-tau Yih, Noah~A Smith, Luke Zettlemoyer, and Tao Yu. 2023.
\newblock One embedder, any task: Instruction-finetuned text embeddings.
\newblock In \emph{Findings of the Association for Computational Linguistics: ACL 2023}, pages 1102--1121.

\bibitem[{Thakur et~al.(2021)Thakur, Reimers, R{\"u}ckl{\'e}, Srivastava, and Gurevych}]{thakur2021beir}
Nandan Thakur, Nils Reimers, Andreas R{\"u}ckl{\'e}, Abhishek Srivastava, and Iryna Gurevych. 2021.
\newblock Beir: A heterogenous benchmark for zero-shot evaluation of information retrieval models.
\newblock \emph{arXiv preprint arXiv:2104.08663}.

\bibitem[{Thorne et~al.(2018)Thorne, Vlachos, Christodoulopoulos, and Mittal}]{thorne2018fever}
James Thorne, Andreas Vlachos, Christos Christodoulopoulos, and Arpit Mittal. 2018.
\newblock Fever: a large-scale dataset for fact extraction and verification.
\newblock In \emph{Proceedings of the 2018 Conference of the North American Chapter of the Association for Computational Linguistics: Human Language Technologies, Volume 1 (Long Papers)}. Association for Computational Linguistics.

\bibitem[{Touvron et~al.(2023)Touvron, Martin, Stone, Albert, Almahairi, Babaei, Bashlykov, Batra, Bhargava, Bhosale et~al.}]{touvron2023llama}
Hugo Touvron, Louis Martin, Kevin Stone, Peter Albert, Amjad Almahairi, Yasmine Babaei, Nikolay Bashlykov, Soumya Batra, Prajjwal Bhargava, Shruti Bhosale, and 1 others. 2023.
\newblock Llama 2: Open foundation and fine-tuned chat models.
\newblock \emph{arXiv preprint arXiv:2307.09288}.

\bibitem[{Wang et~al.(2022)Wang, Yang, Huang, Jiao, Yang, Jiang, Majumder, and Wei}]{wang2022text}
Liang Wang, Nan Yang, Xiaolong Huang, Binxing Jiao, Linjun Yang, Daxin Jiang, Rangan Majumder, and Furu Wei. 2022.
\newblock Text embeddings by weakly-supervised contrastive pre-training.
\newblock \emph{arXiv preprint arXiv:2212.03533}.

\bibitem[{Wang et~al.(2023)Wang, Yang, Huang, Jiao, Yang, Jiang, Majumder, and Wei}]{wang-etal-2023-simlm}
Liang Wang, Nan Yang, Xiaolong Huang, Binxing Jiao, Linjun Yang, Daxin Jiang, Rangan Majumder, and Furu Wei. 2023.
\newblock \href {https://doi.org/10.18653/v1/2023.acl-long.125} {{S}im{LM}: Pre-training with representation bottleneck for dense passage retrieval}.
\newblock In \emph{Proceedings of the 61st Annual Meeting of the Association for Computational Linguistics (Volume 1: Long Papers)}, pages 2244--2258, Toronto, Canada. Association for Computational Linguistics.

\bibitem[{Wang et~al.(2024)Wang, Yang, Huang, Yang, Majumder, and Wei}]{wang2024improving}
Liang Wang, Nan Yang, Xiaolong Huang, Linjun Yang, Rangan Majumder, and Furu Wei. 2024.
\newblock Improving text embeddings with large language models.
\newblock In \emph{Proceedings of the 62nd Annual Meeting of the Association for Computational Linguistics (Volume 1: Long Papers)}, pages 11897--11916.

\bibitem[{Xiao et~al.(2022)Xiao, Liu, Shao, and Cao}]{xiao2022retromae}
Shitao Xiao, Zheng Liu, Yingxia Shao, and Zhao Cao. 2022.
\newblock Retromae: Pre-training retrieval-oriented language models via masked auto-encoder.
\newblock In \emph{Proceedings of the 2022 Conference on Empirical Methods in Natural Language Processing}, pages 538--548.

\bibitem[{Xiao et~al.(2024)Xiao, Liu, Zhang, Muennighoff, Lian, and Nie}]{xiao2024c}
Shitao Xiao, Zheng Liu, Peitian Zhang, Niklas Muennighoff, Defu Lian, and Jian-Yun Nie. 2024.
\newblock C-pack: Packed resources for general chinese embeddings.
\newblock In \emph{Proceedings of the 47th international ACM SIGIR conference on research and development in information retrieval}, pages 641--649.

\bibitem[{Xie et~al.(2023)Xie, Dong, Wang, Lv, Yao, Gan, Wu, Li, Li, Liu et~al.}]{xie2023t2ranking}
Xiaohui Xie, Qian Dong, Bingning Wang, Feiyang Lv, Ting Yao, Weinan Gan, Zhijing Wu, Xiangsheng Li, Haitao Li, Yiqun Liu, and 1 others. 2023.
\newblock T2ranking: A large-scale chinese benchmark for passage ranking.
\newblock In \emph{Proceedings of the 46th International ACM SIGIR Conference on Research and Development in Information Retrieval}, pages 2681--2690.

\bibitem[{Yang et~al.(2018)Yang, Qi, Zhang, Bengio, Cohen, Salakhutdinov, and Manning}]{yang2018hotpotqa}
Zhilin Yang, Peng Qi, Saizheng Zhang, Yoshua Bengio, William Cohen, Ruslan Salakhutdinov, and Christopher~D Manning. 2018.
\newblock Hotpotqa: A dataset for diverse, explainable multi-hop question answering.
\newblock In \emph{Proceedings of the 2018 Conference on Empirical Methods in Natural Language Processing}, pages 2369--2380.

\bibitem[{Zhang et~al.(2021)Zhang, Ma, Shi, and Lin}]{zhang2021mr}
Xinyu Zhang, Xueguang Ma, Peng Shi, and Jimmy Lin. 2021.
\newblock Mr. tydi: A multi-lingual benchmark for dense retrieval.
\newblock In \emph{Proceedings of the 1st Workshop on Multilingual Representation Learning}, pages 127--137.

\bibitem[{Zhang et~al.(2023)Zhang, Thakur, Ogundepo, Kamalloo, Alfonso-Hermelo, Li, Liu, Rezagholizadeh, and Lin}]{zhang2023miracl}
Xinyu Zhang, Nandan Thakur, Odunayo Ogundepo, Ehsan Kamalloo, David Alfonso-Hermelo, Xiaoguang Li, Qun Liu, Mehdi Rezagholizadeh, and Jimmy Lin. 2023.
\newblock Miracl: A multilingual retrieval dataset covering 18 diverse languages.
\newblock \emph{Transactions of the Association for Computational Linguistics}, 11:1114--1131.

\end{thebibliography}

\appendix

\section{Training Details}
\subsection{Training Datasets}
\label{sec:e5}
The public portion of the E5 dataset consists of ELI5 (sample ratio 0.1) \citep{fan2019eli5}, HotpotQA \citep{yang2018hotpotqa}, FEVER \citep{thorne2018fever}, MIRACL \citep{zhang2023miracl}, MS-MARCO passage ranking (sample ratio 0.5) and document ranking (sample ratio 0.2) \citep{nguyen2016ms}, NQ \citep{karpukhin2020dense}, NLI \citep{gao2021simcse}, SQuAD \citep{rajpurkar2016squad}, TriviaQA \citep{joshi2017triviaqa}, Quora Duplicate Questions (sample ratio 0.1) \citep{quora-question-pairs}, Mr- TyDi \citep{zhang2021mr}, DuReader \citep{he2018dureader}, and T2Ranking (sample ratio 0.5) \citep{xie2023t2ranking}. 

During Stage I, we use (query, positive) pairs as the training corpus. For fine-tuning in Stage II, we follow the setup of \citet{wang2024improving}, with dataset-specific instructions summarized in Table \ref{tab:ins_train}.

\renewcommand{\arraystretch}{1.1}
\setlength{\tabcolsep}{4pt}
\setlength{\tabcolsep}{4pt}
\begin{table*}[ht]
\centering
\small
\begin{tabular}{ll}
\toprule
\textbf{Dataset} & \textbf{Instruction(s)} \\
\midrule
NLI & Given a premise, retrieve a hypothesis that is entailed by the premise \\
    & Retrieve semantically similar text \\
DuReader & Given a Chinese search query, retrieve web passages that answer the question \\
ELI5 & Provided a user question, retrieve the highest voted answers on Reddit ELI5 forum \\
FEVER & Given a claim, retrieve documents that support or refute the claim \\
HotpotQA & Given a multi-hop question, retrieve documents that can help answer the question \\
MIRACL & Given a question, retrieve Wikipedia passages that answer the question \\
MrTyDi & Given a question, retrieve Wikipedia passages that answer the question \\
MSMARCO Passage & Given a web search query, retrieve relevant passages that answer the query \\
MSMARCO Document & Given a web search query, retrieve relevant documents that answer the query \\
NQ & Given a question, retrieve Wikipedia passages that answer the question \\
QuoraDuplicates & Given a question, retrieve questions that are semantically equivalent to the given question \\
                & Find questions that have the same meaning as the input question \\
SQuAD & Retrieve Wikipedia passages that answer the question \\
T2Ranking & Given a Chinese search query, retrieve web passages that answer the question \\
TriviaQA & Retrieve Wikipedia passages that answer the question \\
\bottomrule
\end{tabular}
\caption{Instructions used for each of the E5 datasets during fine-tuning.}
\label{tab:ins_train}
\end{table*}

\subsection{Training Setup}
\label{sec:train_detail}
We adopt a two-stage training framework, where the model first undergoes full-parameter training with bidirectional reconstruction objectives, followed by parameter-efficient fine-tuning using LoRA. All experiments are performed with a maximum input length of 512 tokens and FlashAttention-2 enabled.

\paragraph{Stage I.}
In training stage I, models are trained for 2000 steps on the E5 dataset using only (query, positive) pairs. The learning rate is set to $4e-5$, with a total batch size of 512 achieved by gradient accumulation. We apply linear warm-up over the first 300 steps and use end-of-sequence (\texttt{[EOS]}) token pooling to obtain sentence embeddings. Gradient checkpointing is enabled to reduce memory usage.

\paragraph{Stage II.}
Fine-tuning in Stage II is performed for 1000 steps with LoRA (rank 16) using a learning rate of $2e-4$ and a total batch size of 512. The same EOS pooling strategy and gradient checkpointing settings are the same as Stage I. Instructions used for each of the E5 dataset during this stage are summarized in Table~\ref{tab:ins_train}.

\subsection{Fine-tuning Loss}
\begin{figure}
\centering
\includegraphics[width=0.5\textwidth]{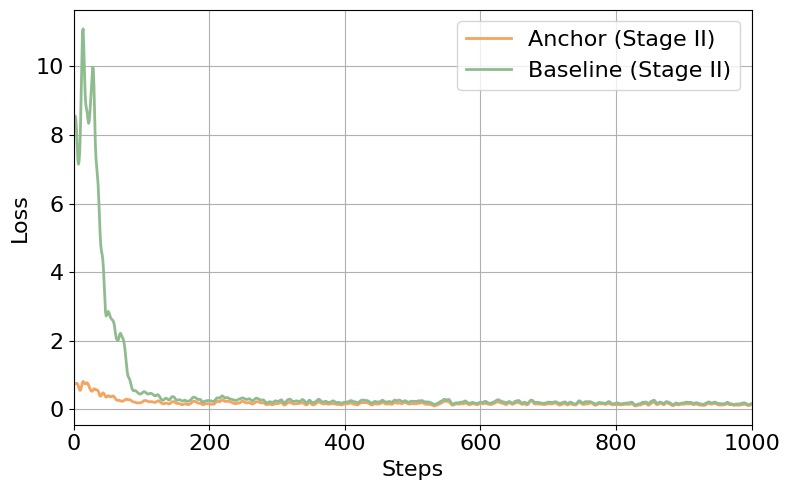}
\caption{Comparison of training loss during fine-tuning: Baseline vs. Training Stage I Initialization (Anchor).}
\label{fig:train_loss}
\end{figure}

Figure~\ref{fig:train_loss} presents the Gaussian-smoothed training loss curves during Stage II fine-tuning of LLaMA-3.2-3B-Instruct. Models that have gone through Stage I (bidirectional reconstruction training) start with significantly lower losses, dropping from over 10 to below 1, compared to models trained directly in Stage II. This demonstrates the effectiveness of our bidirectional reconstruction training. In addition, models trained with Stage I converge faster and more smoothly during Stage II.

\subsection{Training Efficiency of Each Stage}

To provide a clearer view of the computational cost involved in our two-stage training pipeline, we report the training time required for each stage across different model sizes. Specifically, we evaluate the time taken by Stage I (bidirectional reconstruction pretraining) and Stage II (contrastive fine-tuning) on 8×80GB NVIDIA A100 GPUs, with the results summarized in Table~\ref{tab:training-time}.

\begin{table}[t]
\centering
\begin{tabular}{c|cc}
\toprule
\textbf{Model Size} & \textbf{Stage I} & \textbf{Stage II} \\
\hline
1B & $\approx 4 hrs$ & $\approx 2 hrs$ \\
3B & $\approx 12 hrs$ & $\approx5 hrs$ \\
8B & $\approx45 hrs$ & $\approx13 hrs$ \\
\bottomrule
\end{tabular}
\caption{Training time (in hours) for each stage.}
\label{tab:training-time}
\end{table}

\section{Massive Text Embeddings Benchmark (MTEB)}
\subsection{Task Overview}
\label{sec:mteb}
The MTEB benchmark consists of a broad range of embedding tasks, including classification, clustering, pairwise classification, re-ranking, retrieval, sentence similarity (STS), and summarization, aiming to provide a comprehensive and robust evaluation of embedding quality. For evaluation, we follow the instruction templates from \citet{wang2024improving}, as shown in Table \ref{tab:mteb_infer}.

\subsection{MTEB Subset}
\label{sec:mteb_subset}
To speed up the evaluation, we follow the approach of LLM2vec and adopt the same representative subset of 15 tasks from MTEB for our analyses, as shown in Table~\ref{tab:mteb_subset}. This subset was carefully selected to maintain a similar proportional distribution across categories compared to the full MTEB benchmark, ensuring that ablation studies and analyses are not biased toward any specific category or task.

\renewcommand{\arraystretch}{1.1}
\setlength{\tabcolsep}{4pt}
\begin{table}[ht]
\centering
\begin{tabular}{ll}
\toprule
\textbf{Category} & \textbf{Dataset} \\
\midrule
Retrieval (3) & \begin{tabular}[t]{@{}l@{}}
SciFact \\
ArguAna \\
NFCorpus
\end{tabular} \\ \hline
Reranking (2) & \begin{tabular}[t]{@{}l@{}}
StackOverflowDupQ. \\
SciDocsRR
\end{tabular} \\ \hline
Clustering (3) & \begin{tabular}[t]{@{}l@{}}
BiorxivClusteringS2S \\
MedrxivClusteringS2S \\
TwentyNewsgroupsClus.
\end{tabular} \\ \hline
Pair Classification (1) & SprintDuplicateQ. \\ \hline
Classification (3) & \begin{tabular}[t]{@{}l@{}}
Banking77Classification \\
EmotionClassification \\
MassiveIntentClassification
\end{tabular} \\ \hline
STS (3) & \begin{tabular}[t]{@{}l@{}}
STS17 \\
SICK-R \\
STSBenchmark
\end{tabular} \\ \hline
SummEval (0) & - \\ \hline
Overall & 15 datasets \\
\bottomrule
\end{tabular}
\caption{Subset of MTEB tasks for ablation studies.}
\label{tab:mteb_subset}
\end{table}

\section{LLM-Based Baselines}
\paragraph{E5}\cite{wang2024improving} trains text embedding models by fine-tuning open-source decoder-only LLMs using synthetic data generated by proprietary large language models. The synthetic corpus covers a wide range of embedding tasks across 93 languages and is created entirely without human annotation. During training, a standard contrastive loss is applied to learn effective representations from these synthetic text pairs. We only compare E5 results trained on the publicly available portion of the dataset.

\paragraph{GritLM}\cite{muennighoff2024generative} unifies representation and generation training by finetuning a decoder-only LLM on instruction-formatted data for both tasks. It optimizes a contrastive loss for embeddings:

{\small
\[
\mathcal{L}_{\text{Rep}} = -\frac{1}{M} \sum_{i=1}^{M} \log \frac{\exp\left(\tau \cdot \sigma(f_\theta(q^{(i)}), f_\theta(d^{(i)}))\right)}{\sum_{j=1}^{M} \exp\left(\tau \cdot \sigma(f_\theta(q^{(i)}), f_\theta(d^{(j)}))\right)}
\]
}
and a standard language modeling loss for generation:
\[
\mathcal{L}_{\text{Gen}} = -\frac{1}{N} \sum_{i=1}^{N} \log P(f_{\theta,\eta}(x^{(i)}) \mid f_{\theta,\eta}(x^{(<i)})).
\]
The final objective combines both:
\[
\mathcal{L}_{\text{GRIT}} = \lambda_{\text{Rep}} \mathcal{L}_{\text{Rep}} + \lambda_{\text{Gen}} \mathcal{L}_{\text{Gen}}.
\]
This approach enables parameter-efficient training of strong text encoders and generators without modifying model architecture. Similarly, we only compare results trained on publicly available data.

\paragraph{Echo Embeddings}\cite{springer2025repetition} derives high-quality text embeddings from auto-regressive language models without modifying their architecture. The key idea is to repeat the input sequence and extract embeddings from the repeated tokens, which have access to the full context of the original input. This simple repetition strategy enables autoregressive models to approximate bidirectional behavior and significantly improves embedding quality in zero-shot settings. 

\paragraph{LLM2Vec}\cite{behnamghader2024llm2vec} converts decoder-only language models into effective text encoders by enabling bidirectional attention, training with masked next token prediction, and applying contrastive learning. 

\paragraph{bge-en-icl}\cite{li2025making} enhances the text representation ability of decoder-only language models by leveraging in-context learning during training. Instead of relying on task-specific instructions or architectural modifications, it samples a variable number of input examples to simulate in-context scenarios. This strategy equips the model with the ability to generalize across tasks while preserving zero-shot performance. For fair comparison, we only compare results from models trained on the same data and evaluated in zero-shot settings.

\renewcommand{\arraystretch}{1.1}
\setlength{\tabcolsep}{4pt}
\begin{table}[t]
\centering
\begin{tabular}{lccccc}
\toprule
\textbf{Steps} & \textbf{25} & \textbf{50} & \textbf{75} & \textbf{100} & \textbf{125} \\
\midrule

\noalign{\vskip -0.2em}
\multicolumn{6}{c}{ \textbf{LLaMA-3.2-1B-Instruct}} \\
\noalign{\vskip -0.2em}
\midrule
Baseline      & 23.38 & 43.83 & 59.55 & 62.56 & 63.07 \\
OnlyD2Q      & 57.77 & 60.80 & 63.49 & 63.62 & 64.28 \\
OnlyQ2D      & 59.15 & 61.58 & 62.23 & 62.89 & 63.48 \\
\rowcolor[gray]{0.9} Anchor & \textbf{59.31} & \textbf{61.92} & \textbf{63.31} & \textbf{64.40} & \textbf{64.84} \\

\midrule
\noalign{\vskip -0.2em}
\multicolumn{6}{c}{ \textbf{LLaMA-3.2-3B-Instruct}} \\
\noalign{\vskip -0.2em}
\midrule
Baseline      & 36.43 & 48.54 & 58.50 & 63.91 & 64.56 \\
\rowcolor[gray]{0.9} Anchor & \textbf{59.21} & \textbf{62.62} & \textbf{64.03} & \textbf{65.90} & \textbf{66.20} \\

\midrule
\noalign{\vskip -0.2em}
\multicolumn{6}{c}{ \textbf{LLaMA-3.1-8B-Instruct}} \\
\noalign{\vskip -0.2em}
\midrule
Baseline      & 40.35 & 56.79 & 63.78 & 65.29 & 65.39 \\
\rowcolor[gray]{0.9} Anchor & \textbf{61.35} & \textbf{63.79} & \textbf{65.88} & \textbf{67.29} & \textbf{67.36} \\

\bottomrule
\end{tabular}
\caption{Early-stage performance on the MTEB subset during Stage II fine-tuning.}
\label{tab:early-stage}
\end{table}

\section{Experimentals}
\subsection{Detailed Main Results}
We report the detailed performance of baselines and Anchor Embedding models built on LLaMA-3.2-1B-Instruct, Qwen2.5-1.5B-Instruct, LLaMA-3.2-3B-Instruct, and LLaMA-3.1-8B-Instruct across the full MTEB benchmark in Table~\ref{tab:full_rsult1} and Table~\ref{tab:full_result2}. Here, baselines refer to models trained only with Stage II.



\subsection{Early-stage Fine-tuning Results}

In Table~\ref{tab:early-stage}, we report the detailed evaluation scores on the 15-task MTEB subset at checkpoints saved every 25 steps during Stage II fine-tuning. Anchor Embedding consistently outperforms the baselines across all model sizes and converges faster.

\subsection{Results of Training Stage Ablation}  
To support the analysis in Figure~\ref{fig:stage}, we present the exact performance scores of different training stage combinations on the 15-task MTEB subset. As shown in Table~\ref{tab:stage-ablation}, the unsupervised baseline yields the lowest scores, while adding either Stage I (bidirectional reconstruction) or Stage II (contrastive fine-tuning) brings significant gains. The best results are achieved when combining both stages, confirming the effectiveness of our method.

\renewcommand{\arraystretch}{1.1}
\setlength{\tabcolsep}{4pt}
\begin{table}[t]
\centering
\begin{tabular}{l|cc}
\toprule
\textbf{Training Setting} & \textbf{1B} & \textbf{3B} \\
\hline
Unsupervised Baseline & 46.23 & 44.29 \\
Only Stage I (Bi-Reconstruction) & 59.11 & 59.01 \\
Only Stage II (Contrastive) & 63.95 & 65.24 \\
Stage I + Stage II & \textbf{65.19} & \textbf{66.16} \\
\bottomrule
\end{tabular}
\caption{Impact of training stages on performance evaluated on MTEB subset.}
\label{tab:stage-ablation}
\end{table}

\begin{table*}[t]
\centering
\small
\begin{tabular}{lp{5cm}p{7.5cm}}
\toprule
\textbf{Usage} & \textbf{Dataset} & \textbf{License / URL} \\
\hline
Training & Public Portion of E5 \citep{wang2024improving}, curated by \citet{springer2025repetition} & Apache License 2.0 \newline \url{https://github.com/jakespringer/echo-embeddings#training} \newline \url{https://github.com/jakespringer/echo-embeddings/blob/master/LICENSE} \\
\hline
Evaluation & MTEB Benchmark \cite{muennighoff2023mteb} & Apache License 2.0 \newline \url{https://github.com/embeddings-benchmark/mteb} \newline \url{https://github.com/embeddings-benchmark/mteb/blob/main/LICENSE} \\
\bottomrule
\end{tabular}
\caption{License information for datasets used in this work.}
\label{tab:license}
\end{table*}

\section{Comparison with LLaMA2Vec}
To further clarify our method, we provide an extended discussion of the differences between our Anchor and previous related work LLaMA2Vec \citep{li2024llama2vec}.

\subsection{Methodology}

Although both Anchor and LLaMA2Vec adopt reconstruction-based objectives to improve the \texttt{[EOS]} embedding, their formulations and training paradigms differ substantially.

    \paragraph{Reconstruction mechanism.}  
    LLaMA2Vec treats reconstruction as a multi-class classification problem: the \texttt{[EOS]} embedding is directly projected through the LLM’s output head into the vocabulary space to perform token classification. Specifically, the objective function of this problem is derived as:
\[
\min - \frac{1}{|T|} \sum_{t \in T} \log \frac{\exp(e^T W_t)}{\sum_{v \in V} \exp(e^T W_v)}
\]
where \( W \in \mathbb{R}^{d \times |V|} \) is the projection head of LLM; \( V \) indicates the vocabulary space; \( T \) stands for the collection of tokens of the target context (input text itself for \( e_t^\alpha \), the next sentence for \( e_t^\beta \)). 

In contrast, Anchor adopts a bidirectional generative strategy. We reuse the \texttt{[EOS]} embedding from either the query or document as input and apply teacher-forcing language modeling to generate its counterpart sequence (document or query). The training loss is standard cross-entropy over the generated sequence.
    
    \paragraph{Supervision signal.}  
    LLaMA2Vec operates in a fully unsupervised setting, relying only on raw text. Anchor leverages task-relevant supervision from query-document pairs. While this introduces a data dependency, it also brings stronger semantic alignment: the \texttt{[EOS]} embedding acts as a semantic anchor that captures meaningful relationships between queries and documents, enhancing effectiveness for retrieval and re-ranking tasks.

\subsection{Experiment}

\paragraph{Training Efficiency.}  
LLaMA2Vec requires 10{,}000 training steps in its unsupervised adaptation stage (batch size 256). In contrast, Anchor achieves better results with only 2{,}000 steps in Stage~I (batch size 512). Importantly, Stage~I can be trained on the same data used for fine-tuning, without requiring additional annotation.

\paragraph{Effectiveness.}  
We report the performance on the BEIR benchmark using NDCG@10 in Table~\ref{tab:beir_results}. Anchor outperforms LLaMA2Vec while requiring fewer training steps in Stage I, demonstrating both higher efficiency and stronger retrieval effectiveness.


\renewcommand{\arraystretch}{1.1}
\setlength{\tabcolsep}{4pt}
\begin{table}[t]
\centering
\begin{tabular}{lcc}
\toprule
\textbf{Model} & \textbf{Steps (Stage I)} & \textbf{BEIR (N@10)} \\
\midrule
LLaMA2Vec      & 10{,}000 & 56.40 \\
Anchor & 2{,}000 & \textbf{58.07} \\
\bottomrule
\end{tabular}
\caption{Comparison of Anchor and LLaMA2Vec on the BEIR benchmark.}
\label{tab:beir_results}
\end{table}



\section{Related Works}
A number of general-purpose embedding models from industry, such as Alibaba's GTE \cite{li2023towards}, NVIDIA's NV-Embed \cite{lee2024nv}, Tencent's Conan-Embedding \cite{li2024conan}, Google’s Gemini Embedding \cite{lee2025gemini}, and the Jina Embeddings series \cite{gunther2023jina, gunther2023jina2}, have shown strong performance across retrieval and semantic tasks, and are widely used in real-world applications. However, these models are often built with large-scale proprietary data and engineering pipelines that make fair academic comparison difficult. Therefore, we do not directly compare with them, and instead include a brief technical overview.

The method proposed in \citet{li2023towards} introduces a lightweight 110M-parameter Transformer encoder enhanced with rotary positional encodings and gated linear units. It follows a two-stage contrastive training process—unsupervised pre-training on large-scale web corpora followed by supervised fine-tuning on relevance datasets using InfoNCE loss. Despite its compact size, this approach outperforms many larger models on retrieval and classification benchmarks.

NV-Embed \cite{lee2024nv} builds on a 7B decoder-only LLM architecture augmented with a latent-attention pooling layer. During contrastive training, the model discards causal masking, and a two-stage instruction tuning process with curated hard negatives further enhances the learned representations. This yields state-of-the-art results across tasks including semantic retrieval, semantic similarity, reranking, and dense passage retrieval.

Conan-Embedding \cite{li2024conan} uses a 1.4B-parameter encoder trained with dynamic hard negative mining and a cross-GPU balancing loss for scalable negative sampling. The inclusion of LLM-generated prompt–response pairs as weak supervision allows the model to top the Chinese MTEB leaderboard and perform strongly in multilingual scenarios.

Gemini Embedding \cite{lee2025gemini} fine-tunes a multilingual and multimodal LLM to produce 3,000-dimensional embeddings via contrastive learning on high-quality filtered datasets. To better support low-resource languages, it incorporates synthetic data. This design enables strong performance on cross-lingual and cross-modal retrieval benchmarks.

The Jina Embeddings framework \cite{gunther2023jina} employs T5-based encoder-only models, ranging from 35M to 6B parameters. These models are first trained on pairwise contrastive objectives using hundreds of millions of filtered sentence pairs, and then further refined via triplet-margin fine-tuning, with curated hard negatives including negation examples. The resulting embeddings outperform or match much larger models.

Jina Embeddings 2 \cite{gunther2023jina2} adapts a BERT-style encoder with ALiBi and gated linear units to support input lengths up to 8,192 tokens. Its three-stage training—long-sequence masked language modeling, contrastive fine-tuning, and hard-negative refinement—produces embeddings competitive with OpenAI’s ada-002 and establishes new benchmarks for long-document understanding.

\section{URLs and Licenses}

Table~\ref{tab:license} summarizes the license information for the datasets used. All datasets are employed strictly for research purposes and in compliance with their respective licenses and intended usage guidelines.

\renewcommand{\arraystretch}{1.1}
\setlength{\tabcolsep}{4pt}
\setlength{\tabcolsep}{4pt}
\begin{table*}[ht]
\centering
\small
\begin{tabularx}{\linewidth}{@{}lX@{}}
\toprule
\textbf{Task Name} & \textbf{Instruction} \\
\midrule
AmazonCounterfactualClassif. & Classify a given Amazon customer review text as either counterfactual or non-counterfactual \\
AmazonPolarityClassification & Classify Amazon reviews into positive or negative sentiment \\
AmazonReviewsClassification & Classify the given Amazon review into its appropriate rating category \\
Banking77Classification & Given a online banking query, find the corresponding intents \\
EmotionClassification & Classify the emotion expressed in the given Twitter message into one of the six emotions: anger, fear, joy, love, sadness, and surprise \\
ImdbClassification & Classify the sentiment expressed in the given movie review text from the IMDB dataset \\
MassiveIntentClassification & Given a user utterance as query, find the user intents \\
MassiveScenarioClassification & Given a utterance as query, find the user scenarios \\
MTOPDomainClassification & Classify the internet domain of the given utterance in task-oriented conversation \\
MTOPIntentClassification & Classify the intent of the given utterance in task-oriented conversation \\
ToxicConversationsClassif. & Classify the given comments as either toxic or not toxic \\
TweetSentimentClassification & Classify the sentiment of a given tweet as either positive, negative, or neutral \\
ArxivClusteringP2P & Identify the main and secondary category of Arxiv papers based on the titles and abstracts \\
ArxivClusteringS2S & Identify the main and secondary category of Arxiv papers based on the titles \\
BiorxivClusteringP2P & Identify the main category of Biorxiv papers based on the titles and abstracts \\
BiorxivClusteringS2S & Identify the main category of Biorxiv papers based on the titles \\
MedrxivClusteringP2P & Identify the main category of Medrxiv papers based on the titles and abstracts \\
MedrxivClusteringS2S & Identify the main category of Medrxiv papers based on the titles \\
RedditClustering & Identify the topic or theme of Reddit posts based on the titles \\
RedditClusteringP2P & Identify the topic or theme of Reddit posts based on the titles and posts \\
StackExchangeClustering & Identify the topic or theme of StackExchange posts based on the titles \\
StackExchangeClusteringP2P & Identify the topic or theme of StackExchange posts based on the given paragraphs \\
TwentyNewsGroupsClustering & Identify the topic or theme of the given news articles \\
SprintDuplicateQuestions & Retrieve duplicate questions from Sprint forum \\
TwitterSemEval2015 & Retrieve tweets that are semantically similar to the given tweet \\
TwitterURLCorpus & Retrieve tweets that are semantically similar to the given tweet \\
AskUbuntuDupQuestions & Retrieve duplicate questions from AskUbuntu forum \\
MindSmallReranking & Retrieve relevant news articles based on user browsing history \\
SciDocsRR & Given a title of a scientific paper, retrieve the titles of other relevant papers \\
StackOverflowDupQuestions & Retrieve duplicate questions from StackOverflow forum \\
ArguAna & Given a claim, find documents that refute the claim \\
ClimateFEVER & Given a claim about climate change, retrieve documents that support or refute the claim \\
CQADupstackRetrieval & Given a question, retrieve detailed question descriptions from Stackexchange that are duplicates to the given question \\
DBPedia & Given a query, retrieve relevant entity descriptions from DBPedia \\
FEVER & Given a claim, retrieve documents that support or refute the claim \\
FiQA2018 & Given a financial question, retrieve user replies that best answer the question \\
HotpotQA & Given a multi-hop question, retrieve documents that can help answer the question \\
MSMARCO & Given a web search query, retrieve relevant passages that answer the query \\
NFCorpus & Given a question, retrieve relevant documents that best answer the question \\
NQ & Given a question, retrieve Wikipedia passages that answer the question \\
QuoraRetrieval & Given a question, retrieve questions that are semantically equivalent to the given question \\
SCIDOCS & Given a scientific paper title, retrieve paper abstracts that are cited by the given paper \\
SciFact & Given a scientific claim, retrieve documents that support or refute the claim \\
Touche2020 & Given a question, retrieve detailed and persuasive arguments that answer the question \\
TRECCOVID & Given a query on COVID-19, retrieve documents that answer the query \\
STS* & Retrieve semantically similar text \\
BUCC/Tatoeba & Retrieve parallel sentences \\
SummEval & Given a news summary, retrieve other semantically similar summaries \\
\bottomrule
\end{tabularx}
\caption{Instructions used for MTEB evaluation. “STS*” denotes the set of all STS tasks.}
\label{tab:mteb_infer}
\end{table*}

\renewcommand{\arraystretch}{1.1}
\setlength{\tabcolsep}{4pt}
\setlength{\tabcolsep}{4pt}
\begin{table*}[htbp]
\centering
\small
\begin{tabular}{l|cc|cc}
\toprule
\noalign{\vskip -0.2em}
\textbf{Task} & \multicolumn{2}{c|}{\textbf{LLaMA-3.2-1B-Instruct}} & \multicolumn{2}{c}{\textbf{Qwen2.5-1.5B-Instruct}} \\
\noalign{\vskip -0.2em}
              & \textbf{Baseline} & \textbf{Anchor}             & \textbf{Baseline} & \textbf{Anchor} \\
\hline
AmazonCounterfactualClassification & 73.70 & 75.93 & 73.25 & 71.81 \\
AmazonPolarityClassification & 87.11 & 85.81 & 91.51 & 94.32 \\
AmazonReviewsClassification & 44.96 & 43.11 & 45.78 & 48.50 \\
ArguAna & 54.00 & 55.28 & 53.75 & 55.71 \\
ArxivClusteringP2P & 47.09 & 48.05 & 46.00 & 47.93 \\
ArxivClusteringS2S & 42.07 & 41.65 & 40.78 & 39.99 \\
AskUbuntuDupQuestions & 60.42 & 61.73 & 58.22 & 63.31 \\
BIOSSES & 83.91 & 85.87 & 84.67 & 86.14 \\
Banking77Classification & 85.63 & 85.65 & 81.73 & 84.62 \\
BiorxivClusteringP2P & 39.29 & 37.36 & 32.61 & 35.06 \\
BiorxivClusteringS2S & 35.52 & 33.78 & 32.94 & 31.74 \\
CQADupstackRetrieval & 41.04 & 42.78 & 44.79 & 45.28 \\
ClimateFEVER & 33.30 & 34.27 & 34.10 & 32.82 \\
DBPedia & 45.38 & 43.29 & 42.07 & 43.96 \\
EmotionClassification & 49.46 & 50.41 & 50.32 & 48.97 \\
FEVER & 88.48 & 88.48 & 87.85 & 86.61 \\
FiQA2018 & 37.19 & 41.42 & 41.64 & 44.25 \\
HotpotQA & 59.24 & 69.48 & 64.69 & 67.96 \\
ImdbClassification & 72.90 & 76.61 & 76.06 & 88.31 \\
MSMARCO & 38.54 & 39.80 & 38.17 & 39.87 \\
MTOPDomainClassification & 94.02 & 94.51 & 91.55 & 94.52 \\
MTOPIntentClassification & 77.01 & 80.18 & 80.88 & 81.61 \\
MassiveIntentClassification & 76.21 & 74.97 & 77.96 & 76.05 \\
MassiveScenarioClassification & 79.31 & 79.00 & 79.94 & 77.55 \\
MedrxivClusteringP2P & 32.51 & 33.09 & 34.60 & 30.21 \\
MedrxivClusteringS2S & 31.06 & 29.95 & 32.46 & 29.32 \\
MindSmallReranking & 32.01 & 31.61 & 30.11 & 32.26 \\
NFCorpus & 35.98 & 35.87 & 37.29 & 37.15 \\
NQ & 54.62 & 57.66 & 56.52 & 60.31 \\
QuoraRetrieval & 88.22 & 88.83 & 88.46 & 89.17 \\
RedditClustering & 54.23 & 55.89 & 54.07 & 52.27 \\
RedditClusteringP2P & 60.88 & 61.98 & 60.85 & 58.10 \\
SCIDOCS & 18.07 & 20.10 & 20.53 & 20.40 \\
SICK-R & 80.70 & 81.64 & 81.76 & 82.10 \\
STS12 & 71.32 & 73.18 & 76.10 & 76.65 \\
STS13 & 85.02 & 84.36 & 85.72 & 85.12 \\
STS14 & 79.80 & 79.80 & 80.29 & 80.94 \\
STS15 & 84.98 & 85.48 & 87.36 & 87.81 \\
STS16 & 84.49 & 85.68 & 84.24 & 85.27 \\
STS17 & 91.28 & 90.67 & 89.80 & 90.88 \\
STS22 & 65.15 & 67.27 & 64.57 & 66.12 \\
STSBenchmark & 86.08 & 87.01 & 85.67 & 86.38 \\
SciDocsRR & 77.85 & 81.84 & 80.93 & 83.43 \\
SciFact & 68.91 & 74.05 & 71.40 & 74.48 \\
SprintDuplicateQuestions & 88.18 & 95.31 & 94.30 & 94.94 \\
StackExchangeClustering & 65.70 & 66.08 & 63.21 & 67.80 \\
StackExchangeClusteringP2P & 30.86 & 34.42 & 30.44 & 35.23 \\
StackOverflowDupQuestions & 49.49 & 51.40 & 50.18 & 51.53 \\
SummEval & 29.94 & 30.87 & 31.89 & 31.61 \\
TRECCOVID & 70.86 & 78.21 & 74.27 & 83.36 \\
Touche2020 & 17.07 & 19.44 & 19.35 & 23.00 \\
ToxicConversationsClassification & 64.40 & 62.18 & 64.83 & 65.90 \\
TweetSentimentExtractionClassification & 61.34 & 61.24 & 61.79 & 61.99 \\
TwentyNewsgroupsClustering & 48.97 & 50.04 & 45.40 & 47.43 \\
TwitterSemEval2015 & 73.24 & 74.54 & 72.07 & 75.91 \\
TwitterURLCorpus & 86.70 & 86.58 & 82.55 & 86.46 \\
\hline
Average & 60.99 & 62.24 & 61.51 & 62.86 \\
\bottomrule
\end{tabular}
\caption{Results of Anchor Emebeddings and baselines on MTEB.}
\label{tab:full_rsult1}
\end{table*}

\renewcommand{\arraystretch}{1.1}
\setlength{\tabcolsep}{4pt}
\begin{table*}[htbp]
\centering
\small
\begin{tabular}{l|cc|cc}
\toprule
\noalign{\vskip -0.2em}
\textbf{Task} & \multicolumn{2}{c|}{\textbf{LLaMA-3.2-3B-Instruct}} & \multicolumn{2}{c}{\textbf{LLaMA-3.1-8B-Instruct}} \\
\noalign{\vskip -0.2em}
              & \textbf{Baseline} & \textbf{Anchor} & \textbf{Baseline} & \textbf{Anchor} \\
\hline
AmazonCounterfactualClassification & 80.69 & 83.19 & 81.66 & 82.30 \\
AmazonPolarityClassification & 87.65 & 88.19 & 88.84 & 92.26 \\
AmazonReviewsClassification & 46.78 & 47.97 & 46.86 & 47.96 \\
ArguAna & 54.28 & 56.12 & 56.71 & 57.96 \\
ArxivClusteringP2P & 44.71 & 46.75 & 46.10 & 47.99 \\
ArxivClusteringS2S & 39.57 & 43.63 & 44.11 & 44.77 \\
AskUbuntuDupQuestions & 60.63 & 62.63 & 64.18 & 66.87 \\
BIOSSES & 85.79 & 86.22 & 86.98 & 85.00 \\
Banking77Classification & 86.99 & 86.74 & 87.18 & 87.43 \\
BiorxivClusteringP2P & 34.59 & 37.18 & 39.01 & 38.62 \\
BiorxivClusteringS2S & 33.72 & 34.95 & 36.19 & 37.29 \\
CQADupstackRetrieval & 41.38 & 44.20 & 48.37 & 49.22 \\
ClimateFEVER & 34.09 & 33.86 & 35.17 & 37.79 \\
DBPedia & 41.54 & 44.80 & 49.01 & 51.88 \\
EmotionClassification & 49.64 & 50.91 & 50.64 & 51.78 \\
FEVER & 88.55 & 90.82 & 89.12 & 91.68 \\
FiQA2018 & 40.60 & 47.51 & 47.84 & 48.40 \\
HotpotQA & 68.51 & 65.37 & 75.22 & 77.12 \\
ImdbClassification & 80.11 & 81.31 & 82.69 & 84.22 \\
MSMARCO & 39.38 & 41.08 & 41.21 & 43.29 \\
MTOPDomainClassification & 95.02 & 95.81 & 92.15 & 94.57 \\
MTOPIntentClassification & 82.01 & 83.48 & 80.25 & 83.47 \\
MassiveIntentClassification & 77.70 & 78.36 & 79.06 & 78.72 \\
MassiveScenarioClassification & 79.89 & 80.89 & 81.12 & 80.99 \\
MedrxivClusteringP2P & 29.17 & 31.33 & 33.86 & 31.90 \\
MedrxivClusteringS2S & 27.56 & 29.75 & 31.72 & 31.14 \\
MindSmallReranking & 29.58 & 32.31 & 33.53 & 36.01 \\
NFCorpus & 38.47 & 36.62 & 39.82 & 39.82 \\
NQ & 57.80 & 62.30 & 62.16 & 64.16 \\
QuoraRetrieval & 88.53 & 89.18 & 89.80 & 88.85 \\
RedditClustering & 56.86 & 59.45 & 59.89 & 60.02 \\
RedditClusteringP2P & 62.35 & 64.07 & 61.85 & 63.86 \\
SCIDOCS & 18.18 & 17.78 & 21.97 & 23.12 \\
SICK-R & 82.37 & 82.57 & 81.22 & 83.38 \\
STS12 & 76.50 & 76.81 & 76.84 & 76.98 \\
STS13 & 85.53 & 83.93 & 83.42 & 86.81 \\
STS14 & 81.56 & 81.18 & 81.63 & 83.27 \\
STS15 & 87.90 & 85.84 & 87.85 & 88.33 \\
STS16 & 85.86 & 84.63 & 86.31 & 86.72 \\
STS17 & 89.99 & 91.36 & 91.82 & 91.34 \\
STS22 & 65.29 & 65.57 & 67.15 & 67.25 \\
STSBenchmark & 86.49 & 86.69 & 87.78 & 88.55 \\
SciDocsRR & 82.98 & 83.17 & 84.87 & 86.93 \\
SciFact & 74.09 & 75.47 & 76.97 & 79.12 \\
SprintDuplicateQuestions & 96.29 & 95.69 & 94.10 & 96.51 \\
StackExchangeClustering & 69.87 & 69.41 & 68.04 & 68.35 \\
StackExchangeClusteringP2P & 32.38 & 32.72 & 31.26 & 31.22 \\
StackOverflowDupQuestions & 51.33 & 52.97 & 53.23 & 55.72 \\
SummEval & 30.98 & 30.98 & 29.67 & 30.13 \\
TRECCOVID & 71.45 & 80.46 & 79.30 & 81.02 \\
Touche2020 & 18.89 & 19.29 & 17.96 & 22.85 \\
ToxicConversationsClassification & 67.36 & 68.07 & 64.96 & 68.03 \\
TweetSentimentExtractionClassification & 62.16 & 62.17 & 62.27 & 62.32 \\
TwentyNewsgroupsClustering & 46.64 & 51.07 & 50.55 & 51.18 \\
TwitterSemEval2015 & 75.85 & 77.37 & 79.47 & 80.28 \\
TwitterURLCorpus & 86.19 & 86.52 & 86.83 & 89.98 \\
\hline
Average & 62.33 & 63.55 & 64.06 & 65.30 \\
\bottomrule
\end{tabular}
\caption{Results of Anchor Emebeddings and baselines on MTEB.}
\label{tab:full_result2}
\end{table*}

\end{document}